\pdfoutput=1

\documentclass[11pt]{article}

\usepackage[final]{acl}

\usepackage{times}
\usepackage{latexsym}

\usepackage[T1]{fontenc}
\usepackage[utf8]{inputenc}

\usepackage{microtype}

\usepackage{inconsolata}

\usepackage{graphicx}

\usepackage{amsmath}
\usepackage{array}
\usepackage{subcaption}
\usepackage{booktabs}
\usepackage{pifont}

\usepackage{pgfplots}
\usepackage{pgfplotstable}
\pgfplotsset{compat=1.18} 

\definecolor{c0}{HTML}{1f77b4}
\definecolor{c1}{HTML}{ff7f0e}
\definecolor{c2}{HTML}{2ca02c} %
\definecolor{c3}{HTML}{d62728} %
\definecolor{c4}{HTML}{9467bd} 
\definecolor{c5}{HTML}{8c564b}
\definecolor{c6}{HTML}{e377c2}
\definecolor{c7}{HTML}{7f7f7f}
\definecolor{c8}{HTML}{bcbd22}
\definecolor{c9}{HTML}{17becf}

\definecolor{accuracy_color}{HTML}{17becf}
\definecolor{confidence_color}{HTML}{e377c2}

\usepackage{rotating}

\def\addlegendimage{\csname pgfplots@addlegendimage\endcsname}

\title{Know What You do Not Know: Verbalized Uncertainty Estimation Robustness on Corrupted Images in Vision-Language Models}

\author{Mirko Borszukovszki \and Ivo Pascal de Jong \and Matias Valdenegro-Toro \\
        Department of Artificial Intelligence, Bernoulli Institute  \\ University of Groningen, The Netherlands\\
        Email: \texttt{ivo.de.jong@rug.nl}, \texttt{m.a.valdenegro.toro@rug.nl}\\
        }

\begin{document}
\pgfplotstableread[header=has colnames, col sep=comma]{
		level,ece,model,corruption
		0.0,14.674220963172806,claude,gaussian_noise
		1.0,27.1671388101983,claude,gaussian_noise
		2.0,27.36543909348442,claude,gaussian_noise
		3.0,33.31444759206799,claude,gaussian_noise
		4.0,46.9971671388102,claude,gaussian_noise
		5.0,71.38810198300284,claude,gaussian_noise
	}\dataEasyVQAClaudeGaussianNoise
	
	\pgfplotstableread[header=has colnames, col sep=comma]{
		level,ece,model,corruption
		0.0,18.441926345609065,gemini,gaussian_noise
		1.0,18.243626062322946,gemini,gaussian_noise
		2.0,21.41643059490085,gemini,gaussian_noise
		3.0,27.762039660056658,gemini,gaussian_noise
		4.0,35.4957507082153,gemini,gaussian_noise
		5.0,39.6600566572238,gemini,gaussian_noise		
	}\dataEasyVQAGeminiGaussianNoise
	
	\pgfplotstableread[header=has colnames, col sep=comma]{
		level,ece,model,corruption
		0.0,17.25212464589235,gpt,gaussian_noise
		1.0,11.303116147308781,gpt,gaussian_noise
		2.0,13.484419263456092,gpt,gaussian_noise
		3.0,13.087818696883852,gpt,gaussian_noise
		4.0,28.555240793201133,gpt,gaussian_noise
		5.0,28.951841359773372,gpt,gaussian_noise				
	}\dataEasyVQAGPTGaussianNoise

    \pgfplotstableread[header=has colnames, col sep=comma]{
		level,ece,model,corruption
		0.0,14.674220963172806,claude,defocus_blur
		1.0,27.762039660056658,claude,defocus_blur
		2.0,17.45042492917847,claude,defocus_blur
		3.0,37.08215297450425,claude,defocus_blur
		4.0,32.32294617563739,claude,defocus_blur
		5.0,32.91784702549575,claude,defocus_blur		
	}\dataEasyVQAClaudeDefocusBlur
	
	\pgfplotstableread[header=has colnames, col sep=comma]{
		level,ece,model,corruption
		0.0,18.640226628895185,gemini,defocus_blur
		1.0,19.036827195467424,gemini,defocus_blur
		2.0,29.745042492917847,gemini,defocus_blur
		3.0,23.994334277620396,gemini,defocus_blur
		4.0,21.813031161473088,gemini,defocus_blur
		5.0,22.804532577903682,gemini,defocus_blur				
	}\dataEasyVQAGeminiDefocusBlur
	
	\pgfplotstableread[header=has colnames, col sep=comma]{
		level,ece,model,corruption
		0.0,17.25212464589235,gpt,defocus_blur
		1.0,21.61473087818697,gpt,defocus_blur
		2.0,16.657223796033996,gpt,defocus_blur
		3.0,22.209631728045327,gpt,defocus_blur
		4.0,20.424929178470254,gpt,defocus_blur
		5.0,21.41643059490085,gpt,defocus_blur		
	}\dataEasyVQAGPTDefocusBlur

    \pgfplotstableread[header=has colnames, col sep=comma]{
		level,ece,model,corruption
		0.0,14.674220963172806,claude,jpeg_compression
		1.0,34.70254957507082,claude,jpeg_compression
		2.0,16.657223796033996,claude,jpeg_compression
		3.0,24.98583569405099,claude,jpeg_compression
		4.0,30.73654390934844,claude,jpeg_compression
		5.0,29.15014164305949,claude,jpeg_compression		
	}\dataEasyVQAClaudeJPEGCompression
	
	\pgfplotstableread[header=has colnames, col sep=comma]{
		level,ece,model,corruption
		0.0,18.441926345609065,gemini,jpeg_compression
		1.0,26.373937677053824,gemini,jpeg_compression
		2.0,31.52974504249292,gemini,jpeg_compression
		3.0,29.34844192634561,gemini,jpeg_compression
		4.0,28.951841359773372,gemini,jpeg_compression
		5.0,31.3314447592068,gemini,jpeg_compression		
	}\dataEasyVQAGeminiJPEGCompression
	
	\pgfplotstableread[header=has colnames, col sep=comma]{
		level,ece,model,corruption
		0.0,17.25212464589235,gpt,jpeg_compression
		1.0,22.407932011331447,gpt,jpeg_compression
		2.0,13.484419263456092,gpt,jpeg_compression
		3.0,19.8300283286119,gpt,jpeg_compression
		4.0,19.23512747875354,gpt,jpeg_compression
		5.0,32.91784702549575,gpt,jpeg_compression		
	}\dataEasyVQAGPTJPEGCompression

    \pgfplotstableread[header=has colnames, col sep=comma]{
		level,ece,model,corruption
		0,16.676217765042978,claude,gaussian_noise
        1,27.67908309455587,claude,gaussian_noise
        2,14.95702005730659,claude,gaussian_noise
        3,37.306590257879655,claude,gaussian_noise
        4,41.088825214899714,claude,gaussian_noise
        5,63.0945558739255,claude,gaussian_noise
    }\dataHardVQAClaudeGaussianNoise
	
	\pgfplotstableread[header=has colnames, col sep=comma]{
		level,ece,model,corruption
		0,15.988538681948423,gemini,gaussian_noise
        1,27.67908309455587,gemini,gaussian_noise
        2,32.83667621776504,gemini,gaussian_noise
        3,34.212034383954155,gemini,gaussian_noise
        4,42.29226361031518,gemini,gaussian_noise
        5,56.21776504297994,gemini,gaussian_noise
	}\dataHardVQAGeminiGaussianNoise
	
	\pgfplotstableread[header=has colnames, col sep=comma]{
		level,ece,model,corruption
		0, 14.441260744985673,gpt,gaussian_noise
        1, 11.69054441260745,gpt,gaussian_noise 
        2, 20.114613180515757,gpt,gaussian_noise
        3, 24.75644699140401,gpt,gaussian_noise 
        4, 44.35530085959885,gpt,gaussian_noise 
        5, 57.42120343839541,gpt,gaussian_noise				
	}\dataHardVQAGPTGaussianNoise

    \pgfplotstableread[header=has colnames, col sep=comma]{
		level,ece,model,corruption
		0,16.50429799426934,claude,defocus_blur
        1,28.36676217765043,claude,defocus_blur
        2,27.851002865329512,claude,defocus_blur
        3,44.52722063037249,claude,defocus_blur
        4,46.07449856733524,claude,defocus_blur
        5,56.90544412607449,claude,defocus_blur
	}\dataHardVQAClaudeDefocusBlur
	
	\pgfplotstableread[header=has colnames, col sep=comma]{
		level,ece,model,corruption
		0,16.16045845272206,gemini,defocus_blur
        1,26.13180515759312,gemini,defocus_blur
        2,26.13180515759312,gemini,defocus_blur
        3,20.974212034383953,gemini,defocus_blur
        4,35.07163323782235,gemini,defocus_blur
        5,37.47851002865329,gemini,defocus_blur
	}\dataHardVQAGeminiDefocusBlur
	
	\pgfplotstableread[header=has colnames, col sep=comma]{
		level,ece,model,corruption
		0,14.441260744985673,gpt,defocus_blur
        1,15.472779369627506,gpt,defocus_blur
        2,25.100286532951287,gpt,defocus_blur
        3,36.446991404011456,gpt,defocus_blur
        4,22.521489971346703,gpt,defocus_blur
        5,49.34097421203438,gpt,defocus_blur
	}\dataHardVQAGPTDefocusBlur

    \pgfplotstableread[header=has colnames, col sep=comma]{
		level,ece,model,corruption
		0,16.676217765042978,claude,jpeg_compression
        1,20.458452722063036,claude,jpeg_compression
        2,28.02292263610315,claude,jpeg_compression
        3,22.349570200573066,claude,jpeg_compression
        4,33.5243553008596,claude,jpeg_compression
        5,40.05730659025788,claude,jpeg_compression
	}\dataHardVQAClaudeJPEGCompression
	
	\pgfplotstableread[header=has colnames, col sep=comma]{
		level,ece,model,corruption
		0,15.988538681948423,gemini,jpeg_compression
        1,28.02292263610315,gemini,jpeg_compression
        2,30.773638968481375,gemini,jpeg_compression
        3,34.04011461318051,gemini,jpeg_compression
        4,29.5702005730659,gemini,jpeg_compression
        5,37.47851002865329,gemini,jpeg_compression
	}\dataHardVQAGeminiJPEGCompression
	
	\pgfplotstableread[header=has colnames, col sep=comma]{
		level,ece,model,corruption
		0,14.269340974212033,gpt,jpeg_compression
        1,15.472779369627506,gpt,jpeg_compression
        2,19.426934097421203,gpt,jpeg_compression
        3,30.429799426934096,gpt,jpeg_compression
        4,26.303724928366762,gpt,jpeg_compression
        5,26.991404011461317,gpt,jpeg_compression		
	}\dataHardVQAGPTJPEGCompression

    \pgfplotstableread[header=has colnames, col sep=comma]{
		level,acc,model,corruption
		0,15.384615384615385,gpt,gaussian_noise
        1,15.384615384615385,gpt,gaussian_noise
        2,30.76923076923077,gpt,gaussian_noise
        3,7.6923076923076925,gpt,gaussian_noise
        4,7.6923076923076925,gpt,gaussian_noise
        5,16.666666666666664,gpt,gaussian_noise
	}\dataCountingGPTGaussianNoise

    \pgfplotstableread[header=has colnames, col sep=comma]{
		level,acc,model,corruption
		0,15.384615384615385,gemini,gaussian_noise
        1,15.384615384615385,gemini,gaussian_noise
        2,7.6923076923076925,gemini,gaussian_noise
        3,7.6923076923076925,gemini,gaussian_noise
        4,0.0,gemini,gaussian_noise
        5,0.0,gemini,gaussian_noise
	}\dataCountingGeminiGaussianNoise

    \pgfplotstableread[header=has colnames, col sep=comma]{
		level,acc,model,corruption
		0,15.384615384615385,claude,gaussian_noise
        1,15.384615384615385,claude,gaussian_noise
        2,23.076923076923077,claude,gaussian_noise
        3,15.384615384615385,claude,gaussian_noise
        4,9.090909090909092,claude,gaussian_noise
        5,30.76923076923077,claude,gaussian_noise
	}\dataCountingClaudeGaussianNoise

    \pgfplotstableread[header=has colnames, col sep=comma]{
		level,acc,model,corruption
		0,15.384615384615385,gpt,defocus_blur
        1,0.0,gpt,defocus_blur
        2,25.0,gpt,defocus_blur
        3,41.66666666666667,gpt,defocus_blur
        4,30.0,gpt,defocus_blur
        5,20.0,gpt,defocus_blur
	}\dataCountingGPTDefocusBlur

    \pgfplotstableread[header=has colnames, col sep=comma]{
		level,acc,model,corruption
		0,15.384615384615385,gemini,defocus_blur
        1,7.6923076923076925,gemini,defocus_blur
        2,0.0,gemini,defocus_blur
        3,7.6923076923076925,gemini,defocus_blur
        4,0.0,gemini,defocus_blur
        5,0.0,gemini,defocus_blur
	}\dataCountingGeminiDefocusBlur

    \pgfplotstableread[header=has colnames, col sep=comma]{
		level,acc,model,corruption
		0,15.384615384615385,claude,defocus_blur
        1,7.6923076923076925,claude,defocus_blur
        2,7.6923076923076925,claude,defocus_blur
        3,7.6923076923076925,claude,defocus_blur
        4,30.76923076923077,claude,defocus_blur
        5,23.076923076923077,claude,defocus_blur
	}\dataCountingClaudeDefocusBlur

    \pgfplotstableread[header=has colnames, col sep=comma]{
		level,acc,model,corruption
		0,15.384615384615385,gpt,jpeg_compression
        1,15.384615384615385,gpt,jpeg_compression
        2,7.6923076923076925,gpt,jpeg_compression
        3,7.6923076923076925,gpt,jpeg_compression
        4,0.0,gpt,jpeg_compression
        5,15.384615384615385,gpt,jpeg_compression
	}\dataCountingGPTJPEGCompression

    \pgfplotstableread[header=has colnames, col sep=comma]{
		level,acc,model,corruption
		0,15.384615384615385,gemini,jpeg_compression
        1,7.6923076923076925,gemini,jpeg_compression
        2,15.384615384615385,gemini,jpeg_compression
        3,7.6923076923076925,gemini,jpeg_compression
        4,7.6923076923076925,gemini,jpeg_compression
        5,7.6923076923076925,gemini,jpeg_compression
	}\dataCountingGeminiJPEGCompression

    \pgfplotstableread[header=has colnames, col sep=comma]{
		level,acc,model,corruption
		0,15.384615384615385,claude,jpeg_compression
        1,15.384615384615385,claude,jpeg_compression
        2,15.384615384615385,claude,jpeg_compression
        3,7.6923076923076925,claude,jpeg_compression
        4,23.076923076923077,claude,jpeg_compression
        5,7.6923076923076925,claude,jpeg_compression
    }\dataCountingClaudeJPEGCompression

\maketitle

\begin{abstract}
To leverage the full potential of Large Language Models (LLMs) it is crucial to have some information on their answers' uncertainty. This means that the model has to be able to quantify how certain it is in the correctness of a given response. Bad uncertainty estimates can lead to overconfident wrong answers undermining trust in these models. Quite a lot of research has been done on language models that work with text inputs and provide text outputs. Still, since the visual capabilities have been added to these models recently, there has not been much progress on the uncertainty of Visual Language Models (VLMs). We tested three state-of-the-art VLMs on corrupted image data. We found that the severity of the corruption negatively impacted the models' ability to estimate their uncertainty and the models also showed overconfidence in most of the experiments.
\end{abstract}

\section{Introduction}\label{sec:introduction}
\begin{figure}
    \begin{subfigure}[t]{0.48\linewidth}
        \includegraphics[width=\linewidth]{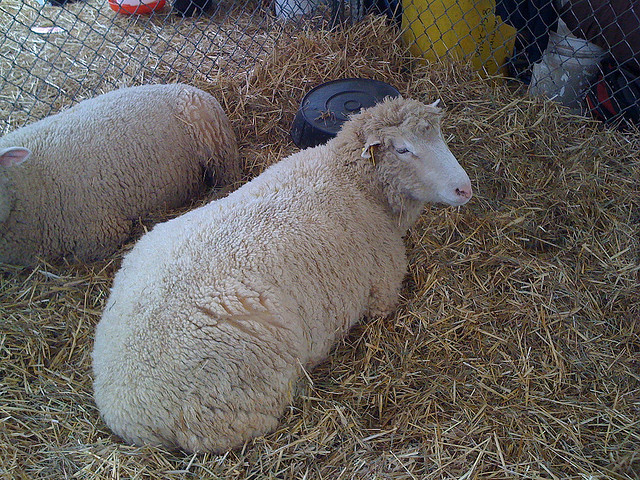}
        \caption{Noise severity 0; \ding{51} \\GPT-4V: "Tag on ear (95\% confidence)" }
    \end{subfigure}%
    \hfill    
    \begin{subfigure}[t]{0.48\linewidth}
        \includegraphics[width=\linewidth]{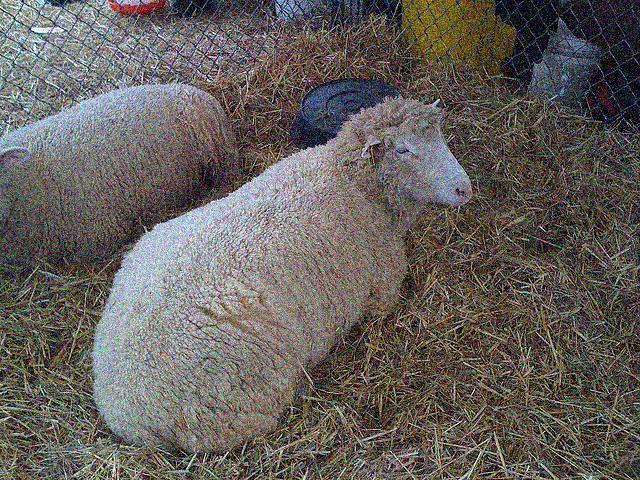}
        \caption{Noise severity 2; \ding{55} \\GPT-4V: "There is nothing unusual (95\% confidence)" }
    \end{subfigure}
    \caption{Question: What is on the sheep? With small noise, GPT-4V is confidently incorrect.}\label{fig:sheep-example}
\end{figure}

LLM-based AI assistants can help us with a wide variety of tasks. The responses generated by these models sound convincing and correct most of the time but it has been shown that they can confidently generate incorrect or even nonsensical answers. In the field of LLMs, this is known as hallucinations \citep{hallucinations}. Currently, the biggest problem with ChatGPT-like AI assistants is that they will generate real and hallucinated answers with the same degree of confidence, as seen in Figure \ref{fig:sheep-example}. As there have already been examples of algorithmic biases with serious consequences in real-world applications of machine learning models \citep{propublicaCOMPAS}, with the rapid evolution of LLMs, it is likely that they will have increasingly more responsibilities in practical applications. There are multiple risks involved with deploying these models in high-stakes decisions in the real world \citep{Ethical_and_social_risks_of_harm_from_language_models, cognitive_bias_hig_stakes}. We have to ensure that these models are well-calibrated, meaning that the model's confidence in a response accurately predicts the likelihood of the answer being correct.

\textbf{Verbalized Uncertainty}. LLMs generate text token by token, from a predefined vocabulary. At each step, the model generates a probability distribution over its vocabulary based on the input and the previously generated tokens and selects the next token from that probability distribution. In theory, the uncertainty of a given answer could be estimated by the combined probability of these tokens \citep{kuhn2023semantic_uncretainty}. Still, since these models are proprietary, we don't have access to these individual token probabilities so methods have been proposed to quantify the uncertainty of a response \citep{tian2023just}. 

To estimate the model's uncertainty in a given answer, we could ask the model in our prompt to quantify it. This is known as verbalized uncertainty \citep{verbalized_uncertinty}. It has been shown that sometimes the models' verbalized confidence estimates are better calibrated than the conditional probabilities estimated via sampling \citep{tian2023just}.

\begin{figure*}[ht]    
    \begin{tabular}{p{0.18\textwidth}p{0.135\textwidth}p{0.135\textwidth}p{0.135\textwidth}p{0.135\textwidth}p{0.135\textwidth}}
      \multicolumn{1}{c}{\textbf{Original Image}}& \multicolumn{1}{c}{\textbf{Severity 1}} & \multicolumn{1}{c}{\textbf{Severity 2}} & \multicolumn{1}{c}{\textbf{Severity 3}} & \multicolumn{1}{c}{\textbf{Severity 4}} & \multicolumn{1}{c}{\textbf{Severity 5}}\\
      \multicolumn{1}{c}{\includegraphics[height=0.85in]{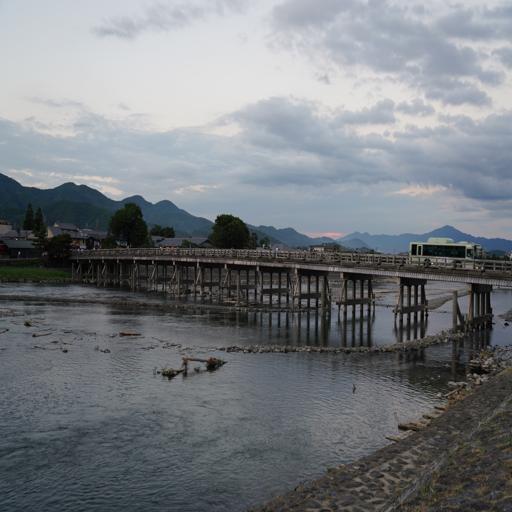}} & \includegraphics[width=\linewidth]{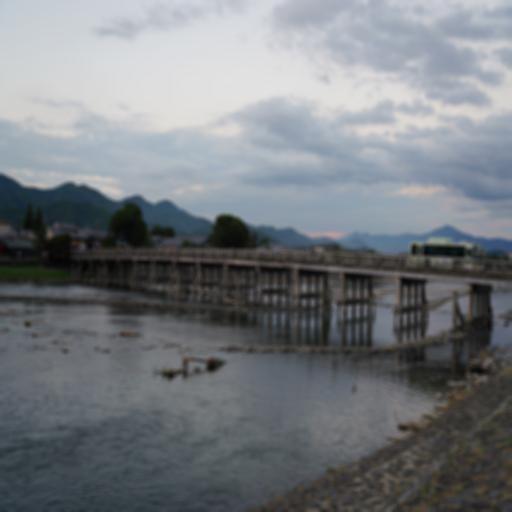} & \includegraphics[width=\linewidth]{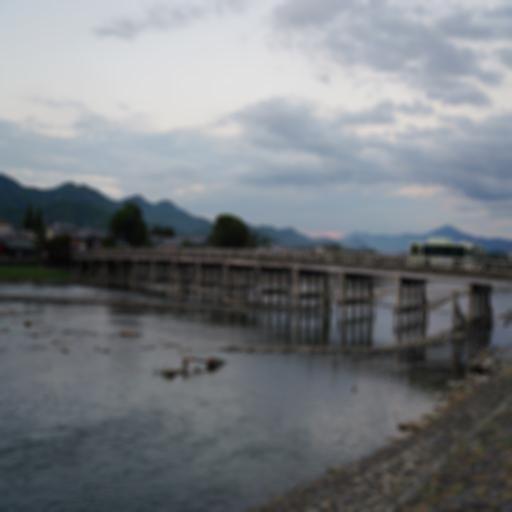} & \includegraphics[width=\linewidth]{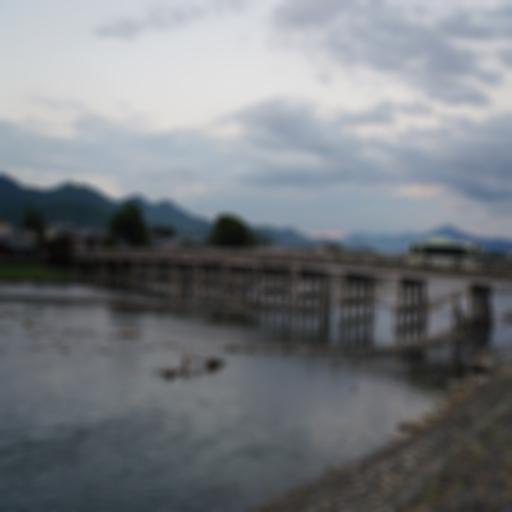} & \includegraphics[width=\linewidth]{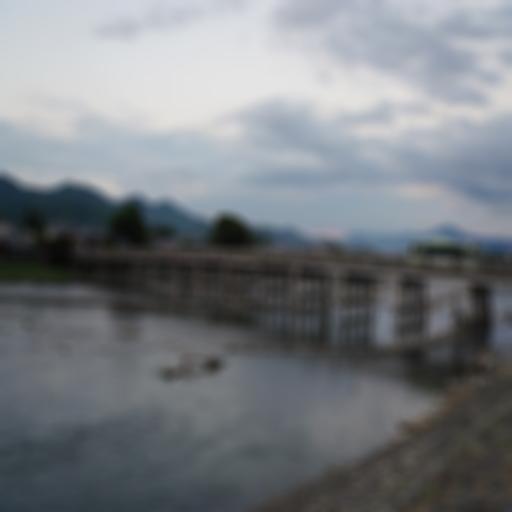} & \includegraphics[width=\linewidth]{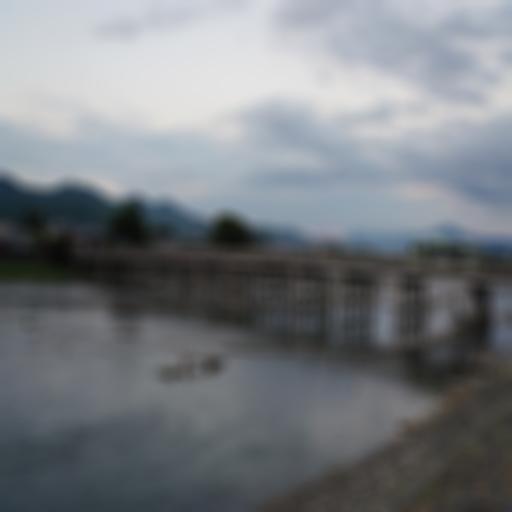}\\
      Togetsukyo Bridge in Arashiyama. & Kyoto, Japan  & Kyoto, Japan & Kyoto, Japan. & Lake Como in Italy. & Lake \\
      \textbf{Conf:} 90\% & \textbf{Conf:} 80\% & \textbf{Conf:} 75\% & \textbf{Conf:} 70\% & \textbf{Conf:} 75\% & \textbf{Conf:} 60\%\\
    \end{tabular}
    \caption{Sample answer from Claude with Defocus Blur Corruption. \textbf{Question}: Where was this photo taken? \textbf{Correct Answer}: Japan, Kyoto, Arashiyama Area, the Bridge is named Togetsu-kyo Bridge (or Toei Bridge). It is clear how answers and confidence degrade with increasing corruption severity. Full answers in Table \ref{tab:claude_answers}.}
    \label{fig:claude_answers_corrupted}
\end{figure*}

Originally, these LLMs could only take in text input and produce text output. However, in the previous five years, multiple advancements were made to extend the capabilities of LLMs to the visual realm. These models can generate text answers from a prompt and an image, or even just an image and are called Visual Language Models (VLMs). One of the first notable examples of these is ViLBERT \citep{vilbert} and two years later CLIP \citep{CLIP}. For a more comprehensive overview of the evolution of VLMs, refer to \citet{evolution_of_vLMs}.  Since then, some of the most widely used LLMs (ChatGPT, Gemini, Claude) have been upgraded with visual understanding. Since they were released in the last two years, there is still much to uncover in understanding their uncertainty.

\begin{figure}[t]
    \centering
    \begin{subfigure}{0.23\linewidth}
        \centering
        \includegraphics[width=\linewidth]{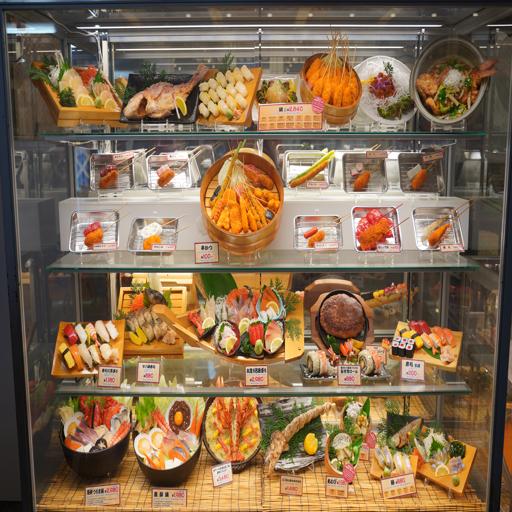}
        \caption{Original\\image}
    \end{subfigure}
    \hfill
    \begin{subfigure}{0.23\linewidth}
        \centering
        \includegraphics[width=\linewidth]{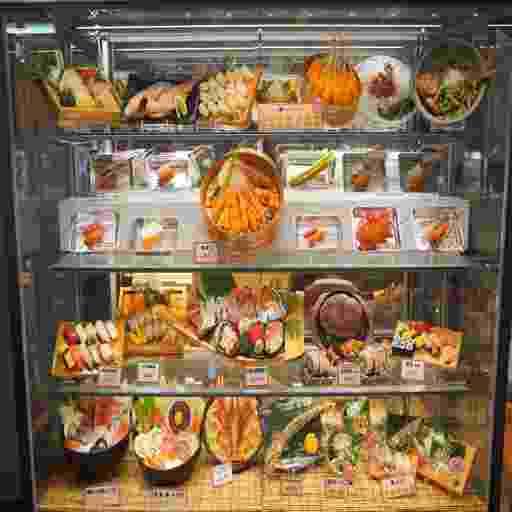}
        \caption{JPEG\\compression}
    \end{subfigure}
    \hfill
    \begin{subfigure}{0.23\linewidth}
        \centering
        \includegraphics[width=\linewidth]{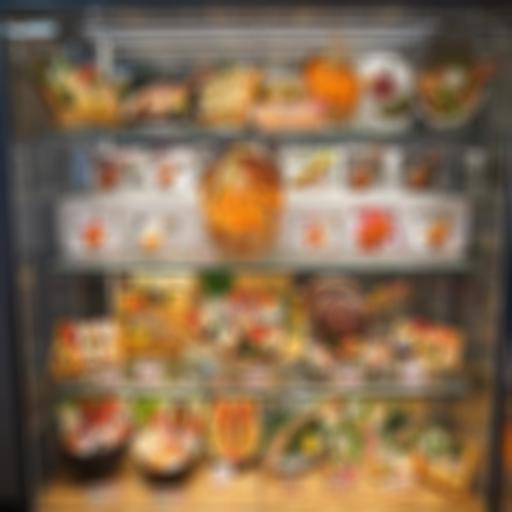}
        \caption{Defocus\\blur}
    \end{subfigure}
    \hfill
    \begin{subfigure}{0.23\linewidth}
        \centering
        \includegraphics[width=\linewidth]{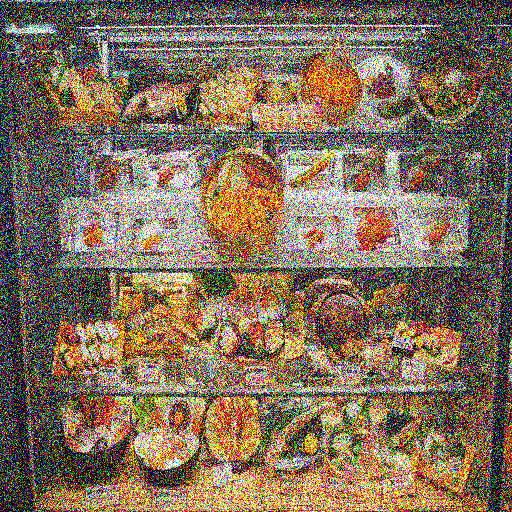}
        \caption{Gaussian\\noise}
    \end{subfigure}
    \caption{Demonstration of the used corruptions on severity 5. \textbf{Question:} What kind of food is showcased in this photo? \textbf{Answer:} Japanese food. Also acceptable is that it is a food model, called Shokuhin Sampuru in Japanese.}
    \label{fig:all_images}
\end{figure}

\textbf{Models and Corruptions.} We tested three state-of-the-art VLMs on visual question-answering tasks where the images are corrupted with common corruptions taken from \citet{corruptions_repo}. It is important to test if a model dealing with image data is robust to these corruptions, as they might not be present in the training set but are likely encountered in a practical application. A demonstration of these corruptions is shown in Figure \ref{fig:all_images}.

There are five severity levels for each corruption each one adding more distortion to the image. This paper aims to answer the research question: \textit{How does the severity level of the corruption impact the model's calibration, accuracy and confidence?} Ideally, as the corruptions become more and more severe and the model starts making mistakes, the confidence should go down along with the accuracy. However, there is evidence that LLMs exhibit overconfidence in their answers \citep{verbalized_uncertinty, overconfidence}, suggesting that increasing severity will increase miscalibration in the models and that the decrease in accuracy will not be accompanied by lower confidence scores.

The three VLMs tested were: GPT-4 Vision \citep{gpt4V}, Gemini Pro Vision \citep{gemini}, and Claude 3 Opus \citep{claude}. We tested all of them on the same image visual question answering tasks where the corruption levels progressively increased. The models were prompted to incorporate their level of uncertainty in their responses or express their answer as a 95\% confidence interval. Figure \ref{fig:claude_answers_corrupted} shows an example answer across corruption severities, with degrading answer quality and decreasing confidence, but still being overconfident.

\section{Related Work}\label{sec:relwork}
In this paper we estimate the model uncertainty by prompting. There is no consensus on the best method to elicit reliable confidence scores from LLMs. This is a problem as different methods yield different confidence scores so it is hard to compare the calibration of different models. \citet{tian2023just} examined various methods to extract confidence scores from the examined models and found that for models trained with Reinforcement Learning with Human Feedback (RLHF) \citep{rlhf}, the verbalized confidence is better calibrated than other methods that for instance, estimate internal token probabilities by sampling. This finding makes verbalized uncertainty a viable option to estimate uncertainty in VLMs.

Even though estimating the model's confidence by prompting has some drawbacks, verbalized uncertainty is getting more attention and has also been examined by \citet{verbalized_uncertinty}. Their work builds on \citet{tian2023just} as they investigate different prompting methods like chain-of-thought reasoning or top-k. Different prompting strategies yielded similar results: LLMs exhibit overconfidence and the majority of the models' confidence scores fall within the 80-100 range. This paper strengthens their findings and tests their "vanilla" prompting strategy on increasingly corrupted images.

Since uncertainty estimation is not often incorporated in computer vision applications \citep{uncertainty_in_CV_valdenegro}, there has not been much research published on the topic. The only paper that examined uncertainty estimation in VLMs is \citet{overconfidence}. They also used verbalized confidence estimation on visual question-answering tasks and found that the models were poorly calibrated, showing severe overconfidence. We build on their research by introducing increasingly corrupted images in the dataset.

Most research \citep{canyoutrust, hendrycks2019benchmarking, rlhftempscale} has been focused on models applied in classification problems or when it comes to question answering, multiple choice or true/false questions. The main issue with this is that their methods for eliciting confidence scores are not applicable to state-of-the-art VLMs. While users would like to enjoy the benefits of well-calibrated models, they should not have to deal with the inner workings of the system and instead receive well-calibrated confidence scores in a verbalized form. We tested the models on more complex, open-ended questions which mimics the usage of these models in the real world. We combined the ideas from \citet{hendrycks2019benchmarking} to test the models on increasingly corrupted images and \citet{overconfidence} to extend the research into VLMs where internal token probabilities are not available. With this paper, we aim to bridge the gap between uncertainty quantification on standard neural networks and VLMs. This is important due to the rapid advancement of VLMs, and the lack of research on their uncertainty calibration.

\section{Methods}\label{sec:methods}

We tested the VLMs on three different datasets using three different corruptions. The specific details of the experiments, datasets, the used corruption techniques and the evaluation procedure are explained below.

\begin{figure}[t]
    \centering
    \begin{subfigure}[b]{0.29\linewidth}
    \centering
        \includegraphics[height=0.9in]{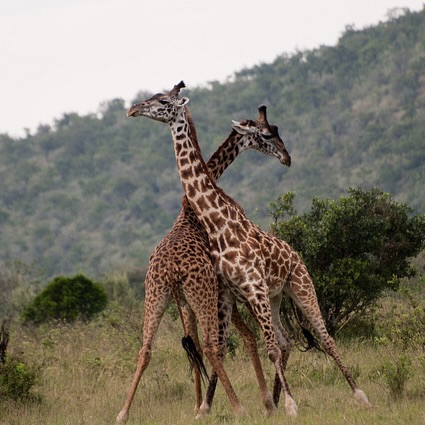}
        \caption{\textbf{Q:} What type of place is this? \mbox{\textbf{A:} Savannah.}}
    \end{subfigure}
    \hfill%
    \begin{subfigure}[b]{0.29\linewidth}
    \centering
        \begin{turn}{90}
        \includegraphics[height=0.9in]{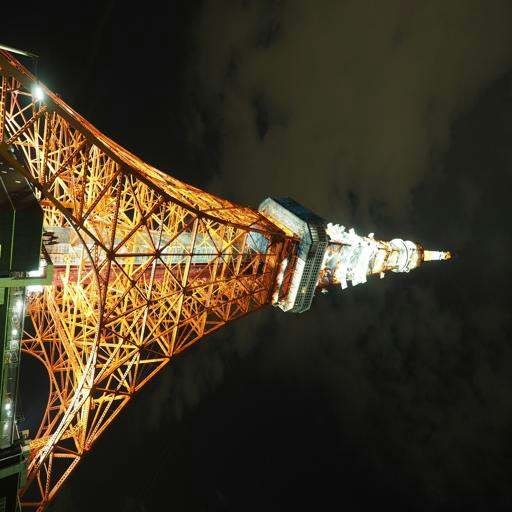}
        \end{turn}
        \caption{\textbf{Q:} What is in this photo? \mbox{\textbf{A:} Tokyo Tower.}}
    \end{subfigure}
    \hfill%
    \begin{subfigure}[b]{0.357\linewidth}
    \centering
        \includegraphics[height=0.9in]{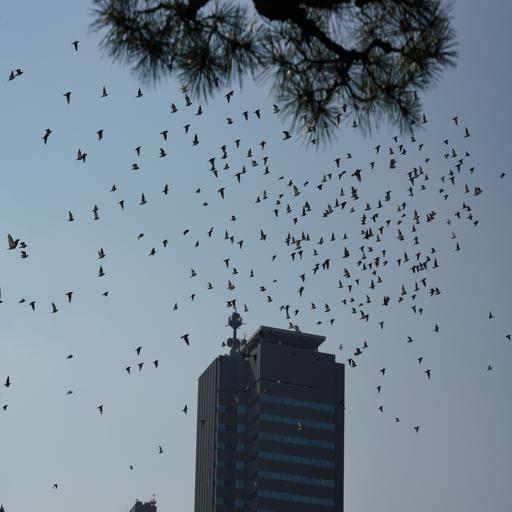}
        \caption{\textbf{Q:} How many birds are shown in this photo? \mbox{\textbf{A:} 250-280.}}
    \end{subfigure}
    \caption{Samples from the three tasks. (a) represents the "easy" task, (b) the "hard" task, (c) the "counting" task. }
    \label{fig:task_demo}
\end{figure}

\subsection{Datasets and data}

The three mentioned models were tested in three experiments:
\begin{enumerate}
    \item \textbf{\emph{Easy} visual question answering} evaluated on the popular visual question answering dataset \citep{VQA, balanced_vqa_v2}. From the testing part of this dataset, 36 randomly sampled images and the corresponding questions were selected. This dataset includes easier questions about images. Without any corruption added to the images, the models should be able to answer most of them.
    \item \textbf{\emph{Hard} visual question answering} evaluated on the Japanese Uncertain Scenes (JUS) dataset proposed by \citet{overconfidence}. This dataset can be downloaded from a public GitHub repository\footnote{https://github.com/ML-RUG/jus-dataset}. This repository contains 29 "tricky" questions specifically designed to evaluate the model's ability to estimate their uncertainty.
    \item The \textbf{Counting task} was also evaluated on the JUS dataset as it contains 13 challenging counting exercises. This is also not designed to evaluate the model's accuracy but rather to check its uncertainty estimates as most of them are nearly impossible to count precisely.
\end{enumerate}

Figure \ref{fig:task_demo} provides example images, questions and answers.
For the selected images and the prompts taken from \citet{overconfidence}, refer to Appendix \ref{ap:appendix}.

\subsection{Experiments}

There were three types of corruption tested and five severity levels for each. Each model was tested on the original dataset and fifteen "corrupted" datasets for each task (3 corruptions, 5 severity levels). Since the models did not always adhere to the requested answer format and there could be multiple equally correct ways to answer an open question, all of the answers had to be manually checked which is the main reason for the low number of images in a particular dataset. Still, this project contains the results of more than 3700 answers across all models and corruptions, counterbalancing the low number of images in a single dataset.

Prompting the models with an image, the question plus the prompt from Appendix \ref{ap:appendix} to elicit verbalized confidence was automated using Python scripts and the APIs provided by OpenAI (GPT-4V), Google (Gemini Pro Vision) and Anthropic (Claude 3 Opus).

For each question, we recorded the confidence score from the model's answer. We also recorded if the answer was correct. Especially at higher severities, there were cases where the image was so distorted that the model refused to respond. For a well-calibrated model, this is a desired behaviour. Because of that, we cannot record that answer as incorrect, but we cannot mark it as correct either as the model did not answer the question. Since in the experiment, we need to measure the models' accuracy, we can only calculate it where each response is marked either correct/incorrect, so in these cases, the answer was not marked as either and no confidence score was recorded.

When the models provided an answer, it was always recorded and used for the analysis, and no data point had to be removed throughout the experiment. However, there were eight cases in the "easy" and "hard" visual question-answering experiments together where Gemini refused to respond due to the image being in conflict with its safety settings. Since there were no explicit images in any of the three datasets, this was most likely due to the model confusing a highly distorted image with explicit content. This confusion was only produced by Gemini and happened only with a small fraction of the tested images.

\subsection{Image Corruptions}
\citet{corruptions_repo} defines 15 types of corruption. They created multiple types of noise and blurring effects and other corruptions mimicking real-life distortions like fog, frost on the lens or snow. These were designed to benchmark neural networks' robustness to corrupted images.

From the 15 corruptions, we investigate three: Gaussian noise, defocus blur and JPEG compression. Different noise-based corruptions have very similar effects so we selected one of them. Gaussian noise or electronic noise is caused by high temperatures or poor lighting conditions \citep{gaussnoise}. Since digital cameras are prone to this type of corruption, the robustness of VLMs against it needs to be tested. Blurring effects like zoom blur or motion blur were discarded as they may change the meaning of picture. Defocus blur does not introduce such ambiguities. Moreover, the most realistic corruption type was chosen from the 15 available ones: JPEG compression. We can safely assume that a VLM encounters images that are distorted due to the lossy nature of the JPEG compression algorithm as these types of digital images are very common.

For a demonstration of the different levels of the three tested corruptions, refer to Appendix \ref{ap:appendixB}.

\subsection{Evaluation Metrics}

Apart from the accuracy and confidence scores, we measured the Expected Calibration Error (ECE) \citep{guo2017calibration}. The formula for calculating the ECE is:
\begin{equation}\label{eq:ece}
  \text{ECE} = \sum_{m=1}^{M}\frac{|B_m|}{n}\left|\text{acc}(B_m) - \text{conf}(B_m)\right|.  
\end{equation}
Where \(M\) is the number of bins, \(|B_m|\) is the number of samples in the \(m\)-th bin, \(n\) is the total number of samples, \(\text{acc}(B_m)\) is the accuracy of the \(m\)-th bin, and \(\text{conf}(B_m)\) is the average confidence of the \(m\)-th bin. This takes the weighted average of the absolute difference between the accuracy and the average confidence of the bins.

This metric quantifies how much one can "trust" the model's confidence scores. The score can be in the range [0, 100] with the ideal ECE of a model being 0, which means that the confidence score accurately predicts the likelihood that the answer is correct. %

\section{Results}\label{sec:results}

Here, we report the results of the three experiments. For visual question answering, we were mainly interested in how the ECE is affected by the increased corruption severity. Since the results seemed to increase linearly, we attempted to fit linear regression lines to the data points and calculated the coefficient of determination $R^2$ value to test the explanatory power of the linear models. %
We were also interested in whether there is any connection between a model's refusal rates and their performance. Especially at higher severity levels, refusing to answer can improve the model's performance. Refusing to answer is not as useful as a correct answer, but it is better than an incorrect guess. In the counting experiment, we only examined the change in accuracy as the corruption severity increased.

\subsection{\emph{Easy} Visual Question Answering}
In this task, the models achieved fairly high accuracy scores on the dataset without any corruption. As the severity of the corruption increased, the models' accuracy started to degrade slightly, but the confidence remained fairly stable. The detailed results are illustrated in Figure \ref{fig:9plot_easy}.

\pgfmathsetlengthmacro\MajorTickLength{
 \pgfkeysvalueof{/pgfplots/major tick length} * 0.5
}
\tikzset{every picture/.style=thick}

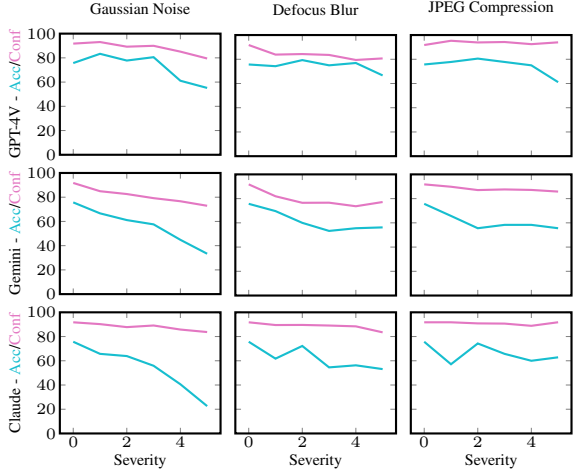
\begin{figure}[t]

    \centering
    \begin{tikzpicture}

        \begin{axis}[ylabel={{\color{black}GPT-4V} - {\color{accuracy_color} Acc}/{\color{confidence_color} Conf}}, ymin=0.0, ymax=100.0, width=0.48\linewidth, label style={font=\tiny}, tick label style={font=\tiny}, major tick length=\MajorTickLength, inner sep=1pt, title={\tiny Gaussian Noise}, xticklabels=\empty]
			\addplot+[confidence_color, no marks] table[x=level, y=metric, col sep=comma] {
				level, metric, model, etc
				0,91.75700934579439,gpt,gaussian_noise,confidence
                1,93.05555555555556,gpt,gaussian_noise,confidence
                2,89.22222222222223,gpt,gaussian_noise,confidence
                3,89.94444444444444,gpt,gaussian_noise,confidence
                4,85.13888888888889,gpt,gaussian_noise,confidence
                5,79.48275862068965,gpt,gaussian_noise,confidence

			};
			\addplot+[accuracy_color, no marks, style={thick}] table[x=level, y=metric, col sep=comma] {
				level, metric, model, etc, type
				0,75.70093457943925,gpt,gaussian_noise,accuracy
                1,83.33333333333334,gpt,gaussian_noise,accuracy
                2,77.77777777777779,gpt,gaussian_noise,accuracy
                3,80.55555555555556,gpt,gaussian_noise,accuracy
                4,61.111111111111114,gpt,gaussian_noise,accuracy
                5,55.172413793103445,gpt,gaussian_noise,accuracy

			};
		\end{axis}
	\end{tikzpicture}
    \begin{tikzpicture}
        \begin{axis}[ymin=0.0, ymax=100.0, width=0.48\linewidth, label style={font=\tiny}, tick label style={font=\tiny}, major tick length=\MajorTickLength, inner sep=1pt, title={\tiny Defocus Blur}, yticklabels=\empty, xticklabels=\empty]
			\addplot+[confidence_color, no marks] table[x=level, y=metric, col sep=comma] {
				level, metric, model, etc
				0,91.75700934579439,gpt,defocus_blur,confidence
                1,83.85714285714286,gpt,defocus_blur,confidence
                2,84.3103448275862,gpt,defocus_blur,confidence
                3,83.57142857142857,gpt,defocus_blur,confidence
                4,79.42307692307692,gpt,defocus_blur,confidence
                5,80.70370370370371,gpt,defocus_blur,confidence
			};
			\addplot+[accuracy_color, no marks] table[x=level, y=metric, col sep=comma] {
				level, metric, model, etc, type
				0,75.70093457943925,gpt,defocus_blur,accuracy
                1,74.28571428571429,gpt,defocus_blur,accuracy
                2,79.3103448275862,gpt,defocus_blur,accuracy
                3,75.0,gpt,defocus_blur,accuracy
                4,76.92307692307693,gpt,defocus_blur,accuracy
                5,66.66666666666666,gpt,defocus_blur,accuracy
			};
		\end{axis}
	\end{tikzpicture}
    \begin{tikzpicture}
        \begin{axis}[ymin=0.0, ymax=100.0, width=0.48\linewidth, label style={font=\tiny}, tick label style={font=\tiny}, major tick length=\MajorTickLength, inner sep=1pt, title={\tiny JPEG Compression}, yticklabels=\empty, xticklabels=\empty]
			\addplot+[confidence_color, no marks] table[x=level, y=metric, col sep=comma] {
				level, metric, model, etc
				0,91.75700934579439,gpt,jpeg_compression,confidence
                1,95.27777777777777,gpt,jpeg_compression,confidence
                2,93.88888888888889,gpt,jpeg_compression,confidence
                3,94.25,gpt,jpeg_compression,confidence
                4,92.47222222222223,gpt,jpeg_compression,confidence
                5,93.97222222222223,gpt,jpeg_compression,confidence
			};
			\addplot+[accuracy_color, no marks] table[x=level, y=metric, col sep=comma] {
				level, metric, model, etc, type
				0,75.70093457943925,gpt,jpeg_compression,accuracy
                1,77.77777777777779,gpt,jpeg_compression,accuracy
                2,80.55555555555556,gpt,jpeg_compression,accuracy
                3,77.77777777777779,gpt,jpeg_compression,accuracy
                4,75.0,gpt,jpeg_compression,accuracy
                5,61.111111111111114,gpt,jpeg_compression,accuracy               
			};
		\end{axis}
	\end{tikzpicture}
    \begin{tikzpicture}
        \begin{axis}[ylabel={{\color{black}Gemini} - {\color{accuracy_color} Acc}/{\color{confidence_color} Conf}}, ymin=0.0, ymax=100.0, width=0.48\linewidth, label style={font=\tiny}, tick label style={font=\tiny}, major tick length=\MajorTickLength, inner sep=1pt, xticklabels=\empty]
			\addplot+[confidence_color, no marks] table[x=level, y=metric, col sep=comma] {
				level, metric, model, etc
                0,91.75700934579439,gemini,gaussian_noise,confidence
                1,84.94444444444444,gemini,gaussian_noise,confidence
                2,82.61111111111111,gemini,gaussian_noise,confidence
                3,79.21212121212122,gemini,gaussian_noise,confidence
                4,76.65517241379311,gemini,gaussian_noise,confidence
                5,72.92592592592592,gemini,gaussian_noise,confidence
			};
			\addplot+[accuracy_color, no marks] table[x=level, y=metric, col sep=comma] {
				level, metric, model, etc, type
				0,75.70093457943925,gemini,gaussian_noise,accuracy
                1,66.66666666666666,gemini,gaussian_noise,accuracy
                2,61.111111111111114,gemini,gaussian_noise,accuracy
                3,57.57575757575758,gemini,gaussian_noise,accuracy
                4,44.827586206896555,gemini,gaussian_noise,accuracy
                5,33.33333333333333,gemini,gaussian_noise,accuracy
			};
		\end{axis}
	\end{tikzpicture}
    \begin{tikzpicture}
        \begin{axis}[ymin=0.0, ymax=100.0, width=0.48\linewidth, label style={font=\tiny}, tick label style={font=\tiny}, major tick length=\MajorTickLength, inner sep=1pt, yticklabels=\empty, xticklabels=\empty]
			\addplot+[confidence_color, no marks] table[x=level, y=metric, col sep=comma] {
				level, metric, model, etc
				0,91.75700934579439,gemini,defocus_blur,confidence
                1,81.93939393939394,gemini,defocus_blur,confidence
                2,76.54285714285714,gemini,defocus_blur,confidence
                3,76.63333333333334,gemini,defocus_blur,confidence
                4,73.70370370370371,gemini,defocus_blur,confidence
                5,77.1875,gemini,defocus_blur,confidence                
			};
			\addplot+[accuracy_color, no marks] table[x=level, y=metric, col sep=comma] {
				level, metric, model, etc, type
				0,75.70093457943925,gemini,defocus_blur,accuracy
                1,69.6969696969697,gemini,defocus_blur,accuracy
                2,60.0,gemini,defocus_blur,accuracy
                3,53.333333333333336,gemini,defocus_blur,accuracy
                4,55.55555555555556,gemini,defocus_blur,accuracy
                5,56.25,gemini,defocus_blur,accuracy                
			};
		\end{axis}
	\end{tikzpicture}
    \begin{tikzpicture}
        \begin{axis}[ymin=0.0, ymax=100.0, width=0.48\linewidth, label style={font=\tiny}, tick label style={font=\tiny}, major tick length=\MajorTickLength, inner sep=1pt, yticklabels=\empty, xticklabels=\empty]
			\addplot+[confidence_color, no marks] table[x=level, y=metric, col sep=comma] {
				level, metric, model, etc
				0,91.75700934579439,gemini,jpeg_compression,confidence
                1,89.8,gemini,jpeg_compression,confidence
                2,87.05555555555556,gemini,jpeg_compression,confidence
                3,87.58333333333333,gemini,jpeg_compression,confidence
                4,87.16666666666667,gemini,jpeg_compression,confidence
                5,85.80555555555556,gemini,jpeg_compression,confidence              
			};
			\addplot+[accuracy_color, no marks] table[x=level, y=metric, col sep=comma] {
				level, metric, model, etc, type
				0,75.70093457943925,gemini,jpeg_compression,accuracy
                1,65.71428571428571,gemini,jpeg_compression,accuracy
                2,55.55555555555556,gemini,jpeg_compression,accuracy
                3,58.333333333333336,gemini,jpeg_compression,accuracy
                4,58.333333333333336,gemini,jpeg_compression,accuracy
                5,55.55555555555556,gemini,jpeg_compression,accuracy
			};
		\end{axis}
	\end{tikzpicture}
    \begin{tikzpicture}
        \begin{axis}[ylabel={{\color{black}Claude} - {\color{accuracy_color} Acc}/{\color{confidence_color} Conf}}, xlabel={Severity}, ymin=0.0, ymax=100.0, width=0.48\linewidth, label style={font=\tiny}, tick label style={font=\tiny}, major tick length=\MajorTickLength, inner sep=1pt]
			\addplot+[confidence_color, no marks] table[x=level, y=metric, col sep=comma] {
				level, metric, model, etc
				0,91.75700934579439,claude,gaussian_noise,confidence
                1,90.25714285714285,claude,gaussian_noise,confidence
                2,87.77777777777777,claude,gaussian_noise,confidence
                3,89.11764705882354,claude,gaussian_noise,confidence
                4,85.78125,claude,gaussian_noise,confidence
                5,83.70967741935483,claude,gaussian_noise,confidence
            };
			\addplot+[accuracy_color, no marks] table[x=level, y=metric, col sep=comma] {
				level, metric, model, etc, type
				0,75.70093457943925,claude,gaussian_noise,accuracy
                1,65.71428571428571,claude,gaussian_noise,accuracy
                2,63.888888888888886,claude,gaussian_noise,accuracy
                3,55.88235294117647,claude,gaussian_noise,accuracy
                4,40.625,claude,gaussian_noise,accuracy
                5,22.58064516129032,claude,gaussian_noise,accuracy
			};
		\end{axis}
	\end{tikzpicture}
    \begin{tikzpicture}
        \begin{axis}[xlabel={Severity}, ymin=0.0, ymax=100.0, width=0.48\linewidth, label style={font=\tiny}, tick label style={font=\tiny}, major tick length=\MajorTickLength, inner sep=1pt, yticklabels=\empty]
			\addplot+[confidence_color, no marks] table[x=level, y=metric, col sep=comma] {
				level, metric, model, etc
				0,91.75700934579439,claude,defocus_blur,confidence
                1,89.5,claude,defocus_blur,confidence
                2,89.58333333333333,claude,defocus_blur,confidence
                3,89.0909090909091,claude,defocus_blur,confidence
                4,88.40625,claude,defocus_blur,confidence
                5,83.46875,claude,defocus_blur,confidence
			};
			\addplot+[accuracy_color, no marks] table[x=level, y=metric, col sep=comma] {
				level, metric, model, etc, type
				0,75.70093457943925,claude,defocus_blur,accuracy
                1,61.76470588235294,claude,defocus_blur,accuracy
                2,72.22222222222221,claude,defocus_blur,accuracy
                3,54.54545454545454,claude,defocus_blur,accuracy
                4,56.25,claude,defocus_blur,accuracy
                5,53.125,claude,defocus_blur,accuracy
			};
		\end{axis}
	\end{tikzpicture}
    \begin{tikzpicture}
        \begin{axis}[xlabel={Severity}, ymin=0.0, ymax=100.0, width=0.48\linewidth, label style={font=\tiny}, tick label style={font=\tiny}, major tick length=\MajorTickLength, inner sep=1pt, yticklabels=\empty]
			\addplot+[confidence_color, no marks] table[x=level, y=metric, col sep=comma] {
				level, metric, model, etc
				0,91.75700934579439,claude,jpeg_compression,confidence
                1,91.77142857142857,claude,jpeg_compression,confidence
                2,90.85714285714286,claude,jpeg_compression,confidence
                3,90.68571428571428,claude,jpeg_compression,confidence
                4,88.85714285714286,claude,jpeg_compression,confidence
                5,91.8,claude,jpeg_compression,confidence
			};
			\addplot+[accuracy_color, no marks] table[x=level, y=metric, col sep=comma] {
				level, metric, model, etc, type
				0,75.70093457943925,claude,jpeg_compression,accuracy
                1,57.14285714285714,claude,jpeg_compression,accuracy
                2,74.28571428571429,claude,jpeg_compression,accuracy
                3,65.71428571428571,claude,jpeg_compression,accuracy
                4,60.0,claude,jpeg_compression,accuracy
                5,62.857142857142854,claude,jpeg_compression,accuracy                
			};
		\end{axis}
	\end{tikzpicture}
    \caption{{\color{accuracy_color}Accuracy} and {\color{confidence_color}confidence} plots for the three examined models and the three corruptions in the \emph{easy} visual question answering experiment.}
    \label{fig:9plot_easy}
\end{figure}

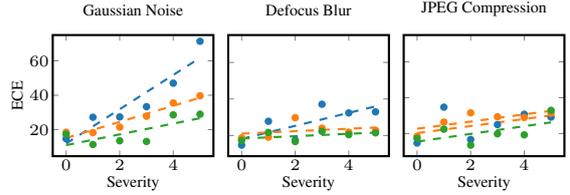
\begin{figure}[t]
    \centering

    \begin{tikzpicture}
		\begin{axis}[ylabel={ECE}, xlabel={Severity}, ymax=75, width=0.48\linewidth, label style={font=\tiny}, tick label style={font=\tiny}, major tick length=\MajorTickLength, inner sep=1pt, title={\tiny Gaussian Noise}]
			\addplot+[c0, only marks, mark size=1pt, mark options={fill=c0}] table[x=level, y=ece] {\dataEasyVQAClaudeGaussianNoise};
			\addplot+[c0, no marks, dashed] table[y={create col/linear regression={x=level, y=ece}}] {\dataEasyVQAClaudeGaussianNoise};
			
			\addplot+[c1, only marks, mark size=1pt, mark options={fill=c1}] table[x=level, y=ece] {\dataEasyVQAGeminiGaussianNoise};
			\addplot+[c1, no marks, dashed] table[y={create col/linear regression={x=level, y=ece}}] {\dataEasyVQAGeminiGaussianNoise};
			
			\addplot+[c2, only marks, mark=*, mark size=1pt, mark options={fill=c2}] table[x=level, y=ece] {\dataEasyVQAGPTGaussianNoise};
			\addplot+[c2, no marks, dashed] table[y={create col/linear regression={x=level, y=ece}}] {\dataEasyVQAGPTGaussianNoise};
		\end{axis}
	\end{tikzpicture}
    \begin{tikzpicture}
		\begin{axis}[xlabel={Severity}, ymax=75, width=0.48\linewidth, label style={font=\tiny}, tick label style={font=\tiny}, major tick length=\MajorTickLength, inner sep=1pt, title={\tiny Defocus Blur}, yticklabels=\empty]
			\addplot+[c0, only marks, mark size=1pt, mark options={fill=c0}] table[x=level, y=ece] {\dataEasyVQAClaudeDefocusBlur};
			\addplot+[c0, no marks, dashed] table[y={create col/linear regression={x=level, y=ece}}] {\dataEasyVQAClaudeDefocusBlur};
			
			\addplot+[c1, only marks, mark size=1pt, mark options={fill=c1}] table[x=level, y=ece] {\dataEasyVQAGeminiDefocusBlur};
			\addplot+[c1, no marks, dashed] table[y={create col/linear regression={x=level, y=ece}}] {\dataEasyVQAGeminiDefocusBlur};
			
			\addplot+[c2, only marks, mark=*, mark size=1pt, mark options={fill=c2}] table[x=level, y=ece] {\dataEasyVQAGPTDefocusBlur};
			\addplot+[c2, no marks, dashed] table[y={create col/linear regression={x=level, y=ece}}] {\dataEasyVQAGPTDefocusBlur};
		\end{axis}
	\end{tikzpicture}
    \begin{tikzpicture}
		\begin{axis}[xlabel={Severity}, ymax=75, width=0.48\linewidth, label style={font=\tiny}, tick label style={font=\tiny}, major tick length=\MajorTickLength, inner sep=1pt, title={\tiny JPEG Compression}, yticklabels=\empty]
			\addplot+[c0, only marks, mark size=1pt, mark options={fill=c0}] table[x=level, y=ece] {\dataEasyVQAClaudeJPEGCompression};
			\addplot+[c1, no marks, dashed] table[y={create col/linear regression={x=level, y=ece}}] {\dataEasyVQAClaudeJPEGCompression};
			
			\addplot+[c1, only marks, mark size=1pt, mark options={fill=c1}] table[x=level, y=ece] {\dataEasyVQAGeminiJPEGCompression};
			\addplot+[c1, no marks, dashed] table[y={create col/linear regression={x=level, y=ece}}] {\dataEasyVQAGeminiJPEGCompression};
			
			\addplot+[c2, only marks, mark=*, mark size=1pt, mark options={fill=c2}] table[x=level, y=ece] {\dataEasyVQAGPTJPEGCompression};
			\addplot+[c2, no marks, dashed] table[y={create col/linear regression={x=level, y=ece}}] {\dataEasyVQAGPTJPEGCompression};
		\end{axis}
	\end{tikzpicture}
    \caption{ECE vs severity level on different corruption types for {\color{c0} Claude}, {\color{c1} Gemini}, and {\color{c2} GPT-4V} in the \emph{easy} visual question answering experiment.    
    }
    \label{fig:ECE_easy}
\end{figure}

We can see that for all models in all corruptions, the average confidence score was higher than the accuracy throughout all severity levels. This means that all models are overconfident. It can also be seen that the gap between the two lines widens as the severity increases. This is not apparent in all plots but is quite visible in the Gaussian noise column. To measure if this gap is actually increasing, we can calculate the ECE scores for each combination of model and corruption type in each severity level and see if it increases as we increase the severity. These ECE scores are visualized in Figure \ref{fig:ECE_easy}

\begin{table}[t]
\resizebox{\linewidth}{!}{%
\begin{tabular}{l c c c }
\toprule
& {\color{c0}Claude} & {\color{c1}Gemini} & {\color{c2}GPT-4V} \\ \midrule
Gaussian Noise & \textbf{0.88} & \textbf{0.93} & 0.53 \\ 
Defocus Blur & 0.54 & 0.11 & 0.28 \\ 
JPEG Compression & 0.21 & 0.58 & 0.36 \\ \bottomrule
\end{tabular}}
\caption{R$^2$ values of ECE vs. severity trends in Fig. \ref{fig:ECE_easy}.}
\label{tab:r_squared_easy}
\end{table}

\begin{figure}[!t] 
    \centering
    \begin{tikzpicture}
		\begin{axis}[ylabel={Refusal Rate}, xlabel={Severity}, ymax=1.0, width=0.48\linewidth, label style={font=\tiny}, tick label style={font=\tiny}, major tick length=\MajorTickLength, inner sep=1pt, title={\tiny Gaussian Noise}]
			\addplot+[c2, no marks] table[x=level, y=refrate, col sep=comma] {
				refrate, level, model, etc
				0.0,0,gpt,gaussian_noise
				0.0,1,gpt,gaussian_noise
				0.0,2,gpt,gaussian_noise
				0.0,3,gpt,gaussian_noise
				0.0,4,gpt,gaussian_noise
				0.194444444444444,5,gpt,gaussian_noise
			};
			\addplot+[c0, no marks] table[x=level, y=refrate, col sep=comma] {
				refrate, level, model, etc
				0.0277777777777778,0,claude,gaussian_noise
				0.0277777777777778,1,claude,gaussian_noise
				0.0,2,claude,gaussian_noise
				0.0555555555555556,3,claude,gaussian_noise
				0.111111111111111,4,claude,gaussian_noise
				0.138888888888889,5,claude,gaussian_noise				
			};
			\addplot+[c1, no marks] table[x=level, y=refrate, col sep=comma] {
				refrate, level, model, etc
				0.0,0,gemini,gaussian_noise
				0.0,1,gemini,gaussian_noise
				0.0,2,gemini,gaussian_noise
				0.0833333333333333,3,gemini,gaussian_noise
				0.194444444444444,4,gemini,gaussian_noise
				0.25,5,gemini,gaussian_noise								
			};
		\end{axis}
	\end{tikzpicture}
    \begin{tikzpicture}
		\begin{axis}[xlabel={Severity}, ymax=1.0, width=0.48\linewidth, label style={font=\tiny}, tick label style={font=\tiny}, major tick length=\MajorTickLength, inner sep=1pt, title={\tiny Defocus Blur}, yticklabels=\empty]]
			\addplot+[c2, no marks] table[x=level, y=refrate, col sep=comma] {
				refrate, level, model, etc
				0.0,0,gpt,defocus_blur
				0.0277777777777778,1,gpt,defocus_blur
				0.194444444444444,2,gpt,defocus_blur
				0.222222222222222,3,gpt,defocus_blur
				0.277777777777778,4,gpt,defocus_blur
				0.25,5,gpt,defocus_blur
				
			};
			\addplot+[c0, no marks] table[x=level, y=refrate, col sep=comma] {
				refrate, level, model, etc
				0.0277777777777778,0,claude,defocus_blur
				0.0555555555555556,1,claude,defocus_blur
				0.0,2,claude,defocus_blur
				0.0833333333333333,3,claude,defocus_blur
				0.111111111111111,4,claude,defocus_blur
				0.111111111111111,5,claude,defocus_blur				
			};
			\addplot+[c1, no marks] table[x=level, y=refrate, col sep=comma] {
				refrate, level, model, etc
				0.0,0,gemini,defocus_blur
				0.0833333333333333,1,gemini,defocus_blur
				0.0277777777777778,2,gemini,defocus_blur
				0.166666666666667,3,gemini,defocus_blur
				0.25,4,gemini,defocus_blur
				0.111111111111111,5,gemini,defocus_blur									
			};	
		\end{axis}
	\end{tikzpicture}
    \begin{tikzpicture}
		\begin{axis}[xlabel={Severity}, ymax=1.0, width=0.48\linewidth, label style={font=\tiny}, tick label style={font=\tiny}, major tick length=\MajorTickLength, inner sep=1pt, title={\tiny JPEG Compression}, yticklabels=\empty]]
			\addplot+[c2, no marks] table[x=level, y=refrate, col sep=comma] {
				refrate, level, model, etc
				0.0,0,gpt,jpeg_compresion
				0.0,1,gpt,jpeg_compresion
				0.0,2,gpt,jpeg_compresion
				0.0,3,gpt,jpeg_compresion
				0.0,4,gpt,jpeg_compresion
				0.0,5,gpt,jpeg_compresion				
			};
			\addplot+[c0, no marks] table[x=level, y=refrate, col sep=comma] {
				refrate, level, model, etc
				0.0277777777777778,0,claude,jpeg_compresion
				0.0277777777777778,1,claude,jpeg_compresion
				0.0277777777777778,2,claude,jpeg_compresion
				0.0277777777777778,3,claude,jpeg_compresion
				0.0277777777777778,4,claude,jpeg_compresion
				0.0277777777777778,5,claude,jpeg_compresion				
			};
			\addplot+[c1, no marks] table[x=level, y=refrate, col sep=comma] {
				refrate, level, model, etc
				0.0,0,gemini,jpeg_compresion
				0.0277777777777778,1,gemini,jpeg_compresion
				0.0,2,gemini,jpeg_compresion
				0.0,3,gemini,jpeg_compresion
				0.0,4,gemini,jpeg_compresion
				0.0,5,gemini,jpeg_compresion				
			};
		\end{axis}
	\end{tikzpicture}
    \caption{Refusal rates for {\color{c0} Claude}, {\color{c1} Gemini}, and {\color{c2} GPT-V4} across different severity levels and corruptions in the \emph{easy} visual question answering experiment.}
    \label{fig:refusals_easy}
\end{figure}

The Expected Calibration Error increases as we increase the severity for all models and corruptions. However, this effect is very small for defocus blur and JPEG compression. The worst performance is shown by Claude 3 Opus with the Gaussian noise corruption. GPT-4V, on the other hand, outperforms the other two models in all three corruptions, achieving the lowest ECE scores across different severity levels.

Table \ref{tab:r_squared_easy} summarizes the R$^2$ values for the linear regression lines presented in Figure \ref{fig:ECE_easy}. The only high values are for Claude and Gemini for Gaussian noise corruption. From Figure \ref{fig:ECE_easy} we can see that these are the two steepest lines in the plot, meaning that increasing severity had the most effect on the model's ECE in these two cases. The high R$^2$ values indicate that the increased severity explains a lot of variance in the ECE. The calibration errors are further investigated with calibration plots in Appendix \ref{ap:calibration_plots}. These show that GPT-4V outperformed the other two models in all types of corruptions.

At higher severity levels, the models sometimes refused to answer and express their confidence score. The refusal rates are summarized in Figure \ref{fig:refusals_easy}.

We can look at the relationship between this plot and Figures \ref{fig:ECE_easy} and \ref{fig:claibration_easy}. All models show similar refusal rates for Gaussian noise but achieve different results in the ECE and the calibration plot. For defocus blur, GPT-4V's refusal rates are much higher than the other two models' and it outperforms them both in ECE scores and the calibration plot. The models showed very low refusal rates and their ECE performance is similar for JPEG compression but there are still small differences between the models in the calibration plot as GPT-4V still shows the best performance. Overall, we have some evidence suggesting that refusing to answer a question can prevent a model from making a wrong prediction, thus improving its accuracy and ECE score, but we see that this is not true for all corruption types.

\subsection{\emph{Hard} Visual Question Answering}

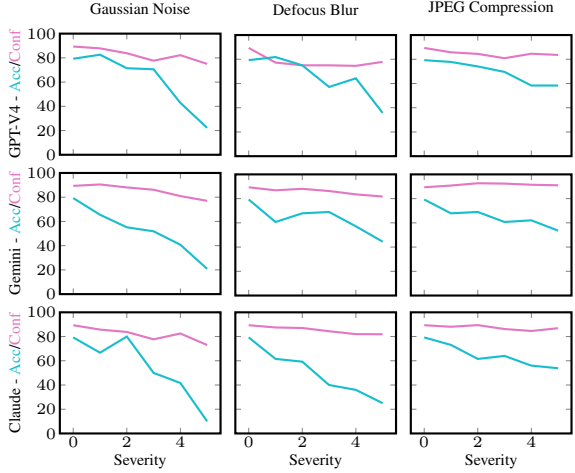
\begin{figure}[t]
    \centering
    \begin{tikzpicture}
        \begin{axis}[ylabel={GPT-V4 - {\color{accuracy_color} Acc}/{\color{confidence_color} Conf}}, ymin=0.0, ymax=100.0, width=0.48\linewidth, label style={font=\tiny}, tick label style={font=\tiny}, major tick length=\MajorTickLength, inner sep=1pt, title={\tiny Gaussian Noise}, xticklabels=\empty]
			\addplot+[confidence_color, no marks] table[x=level, y=metric, col sep=comma] {
				level, metric, model, etc
				0,89.35365853658537,gpt,gaussian_noise,confidence
				1,87.82608695652173,gpt,gaussian_noise,confidence
				2,83.80952380952381,gpt,gaussian_noise,confidence
				3,77.6470588235294,gpt,gaussian_noise,confidence
				4,82.21428571428571,gpt,gaussian_noise,confidence
				5,75.0,gpt,gaussian_noise,confidence
			};
			\addplot+[accuracy_color, no marks] table[x=level, y=metric, col sep=comma] {
				level, metric, model, etc, type
				0,79.26829268292683,gpt,gaussian_noise,accuracy
				1,82.6086956521739,gpt,gaussian_noise,accuracy
				2,71.42857142857143,gpt,gaussian_noise,accuracy
				3,70.58823529411765,gpt,gaussian_noise,accuracy
				4,42.857142857142854,gpt,gaussian_noise,accuracy
				5,22.22222222222222,gpt,gaussian_noise,accuracy				
			};
		\end{axis}
	\end{tikzpicture}
    \begin{tikzpicture}
        \begin{axis}[ymin=0.0, ymax=100.0, width=0.48\linewidth, label style={font=\tiny}, tick label style={font=\tiny}, major tick length=\MajorTickLength, inner sep=1pt, title={\tiny Defocus Blur}, yticklabels=\empty, xticklabels=\empty]
			\addplot+[confidence_color, no marks] table[x=level, y=metric, col sep=comma] {
				level, metric, model, etc
				0,89.35365853658537,gpt,defocus_blur,confidence
                1,77.27272727272727,gpt,defocus_blur,confidence
                2,75.0,gpt,defocus_blur,confidence
                3,75.0,gpt,defocus_blur,confidence
                4,74.57142857142857,gpt,defocus_blur,confidence
                5,77.85714285714286,gpt,defocus_blur,confidence
			};
			\addplot+[accuracy_color, no marks] table[x=level, y=metric, col sep=comma] {
				level, metric, model, etc, type
				0,79.26829268292683,gpt,defocus_blur,accuracy
                1,81.81818181818183,gpt,defocus_blur,accuracy
                2,75.0,gpt,defocus_blur,accuracy
                3,57.14285714285714,gpt,defocus_blur,accuracy
                4,64.28571428571429,gpt,defocus_blur,accuracy
                5,35.714285714285715,gpt,defocus_blur,accuracy
			};
		\end{axis}
	\end{tikzpicture}
    \begin{tikzpicture}
        \begin{axis}[ymin=0.0, ymax=100.0, width=0.48\linewidth, label style={font=\tiny}, tick label style={font=\tiny}, major tick length=\MajorTickLength, inner sep=1pt, title={\tiny JPEG Compression}, yticklabels=\empty, xticklabels=\empty]
			\addplot+[confidence_color, no marks] table[x=level, y=metric, col sep=comma] {
				level, metric, model, etc
				0,89.35365853658537,gpt,jpeg_compression,confidence
                1,85.74074074074075,gpt,jpeg_compression,confidence
                2,84.44444444444444,gpt,jpeg_compression,confidence
                3,80.8695652173913,gpt,jpeg_compression,confidence
                4,84.58333333333333,gpt,jpeg_compression,confidence
                5,83.58333333333333,gpt,jpeg_compression,confidence
			};
			\addplot+[accuracy_color, no marks] table[x=level, y=metric, col sep=comma] {
				level, metric, model, etc, type
				0,79.26829268292683,gpt,jpeg_compression,accuracy
                1,77.77777777777779,gpt,jpeg_compression,accuracy
                2,74.07407407407408,gpt,jpeg_compression,accuracy
                3,69.56521739130434,gpt,jpeg_compression,accuracy
                4,58.333333333333336,gpt,jpeg_compression,accuracy
                5,58.333333333333336,gpt,jpeg_compression,accuracy                
			};
		\end{axis}
	\end{tikzpicture}
    \begin{tikzpicture}
        \begin{axis}[ylabel={Gemini - {\color{accuracy_color} Acc}/{\color{confidence_color} Conf}}, ymin=0.0, ymax=100.0, width=0.48\linewidth, label style={font=\tiny}, tick label style={font=\tiny}, major tick length=\MajorTickLength, inner sep=1pt, xticklabels=\empty]
			\addplot+[confidence_color, no marks] table[x=level, y=metric, col sep=comma] {
				level, metric, model, etc
                0,89.35365853658537,gemini,gaussian_noise,confidence
                1,90.51724137931035,gemini,gaussian_noise,confidence
                2,88.06896551724138,gemini,gaussian_noise,confidence
                3,86.14814814814815,gemini,gaussian_noise,confidence
                4,80.92592592592592,gemini,gaussian_noise,confidence
                5,77.0,gemini,gaussian_noise,confidence                
			};
			\addplot+[accuracy_color, no marks] table[x=level, y=metric, col sep=comma] {
				level, metric, model, etc, type
				0,79.26829268292683,gemini,gaussian_noise,accuracy
                1,65.51724137931035,gemini,gaussian_noise,accuracy
                2,55.172413793103445,gemini,gaussian_noise,accuracy
                3,51.85185185185185,gemini,gaussian_noise,accuracy
                4,40.74074074074074,gemini,gaussian_noise,accuracy
                5,20.833333333333336,gemini,gaussian_noise,accuracy                
			};
		\end{axis}
	\end{tikzpicture}
    \begin{tikzpicture}
        \begin{axis}[ymin=0.0, ymax=100.0, width=0.48\linewidth, label style={font=\tiny}, tick label style={font=\tiny}, major tick length=\MajorTickLength, inner sep=1pt, yticklabels=\empty, xticklabels=\empty]
			\addplot+[confidence_color, no marks] table[x=level, y=metric, col sep=comma] {
				level, metric, model, etc
				0,89.35365853658537,gemini,defocus_blur,confidence
                1,86.85714285714286,gemini,defocus_blur,confidence
                2,88.07142857142857,gemini,defocus_blur,confidence
                3,86.27586206896552,gemini,defocus_blur,confidence
                4,83.5,gemini,defocus_blur,confidence
                5,81.74074074074075,gemini,defocus_blur,confidence                
			};
			\addplot+[accuracy_color, no marks] table[x=level, y=metric, col sep=comma] {
				level, metric, model, etc, type
				0,79.26829268292683,gemini,defocus_blur,accuracy
                1,60.71428571428571,gemini,defocus_blur,accuracy
                2,67.85714285714286,gemini,defocus_blur,accuracy
                3,68.96551724137932,gemini,defocus_blur,accuracy
                4,57.14285714285714,gemini,defocus_blur,accuracy
                5,44.44444444444444,gemini,defocus_blur,accuracy                
			};
		\end{axis}
	\end{tikzpicture}
    \begin{tikzpicture}
        \begin{axis}[ymin=0.0, ymax=100.0, width=0.48\linewidth, label style={font=\tiny}, tick label style={font=\tiny}, major tick length=\MajorTickLength, inner sep=1pt, yticklabels=\empty, xticklabels=\empty]
			\addplot+[confidence_color, no marks] table[x=level, y=metric, col sep=comma] {
				level, metric, model, etc
				0,89.35365853658537,gemini,jpeg_compression,confidence
                1,90.78571428571429,gemini,jpeg_compression,confidence
                2,92.6896551724138,gemini,jpeg_compression,confidence
                3,92.39285714285714,gemini,jpeg_compression,confidence
                4,91.44827586206897,gemini,jpeg_compression,confidence
                5,91.0,gemini,jpeg_compression,confidence                
			};
			\addplot+[accuracy_color, no marks] table[x=level, y=metric, col sep=comma] {
				level, metric, model, etc, type
				0,79.26829268292683,gemini,jpeg_compression,accuracy
                1,67.85714285714286,gemini,jpeg_compression,accuracy
                2,68.96551724137932,gemini,jpeg_compression,accuracy
                3,60.71428571428571,gemini,jpeg_compression,accuracy
                4,62.06896551724138,gemini,jpeg_compression,accuracy
                5,53.57142857142857,gemini,jpeg_compression,accuracy
			};
		\end{axis}
	\end{tikzpicture}
    \begin{tikzpicture}
        \begin{axis}[ylabel={Claude - {\color{accuracy_color} Acc}/{\color{confidence_color} Conf}}, xlabel={Severity}, ymin=0.0, ymax=100.0, width=0.48\linewidth, label style={font=\tiny}, tick label style={font=\tiny}, major tick length=\MajorTickLength, inner sep=1pt]
			\addplot+[confidence_color, no marks] table[x=level, y=metric, col sep=comma] {
				level, metric, model, etc
				0,89.35365853658537,claude,gaussian_noise,confidence
                1,85.74074074074075,claude,gaussian_noise,confidence
                2,83.8,claude,gaussian_noise,confidence
                3,77.6923076923077,claude,gaussian_noise,confidence
                4,82.5,claude,gaussian_noise,confidence
                5,73.0,claude,gaussian_noise,confidence
            };
			\addplot+[accuracy_color, no marks] table[x=level, y=metric, col sep=comma] {
				level, metric, model, etc, type
				0,79.26829268292683,claude,gaussian_noise,accuracy
                1,66.66666666666666,claude,gaussian_noise,accuracy
                2,80.0,claude,gaussian_noise,accuracy
                3,50.0,claude,gaussian_noise,accuracy
                4,41.66666666666667,claude,gaussian_noise,accuracy
                5,10.0,claude,gaussian_noise,accuracy
			};
		\end{axis}
	\end{tikzpicture}
    \begin{tikzpicture}
        \begin{axis}[xlabel={Severity}, ymin=0.0, ymax=100.0, width=0.48\linewidth, label style={font=\tiny}, tick label style={font=\tiny}, major tick length=\MajorTickLength, inner sep=1pt, yticklabels=\empty]
			\addplot+[confidence_color, no marks] table[x=level, y=metric, col sep=comma] {
				level, metric, model, etc
				0,89.35365853658537,claude,defocus_blur,confidence
                1,87.5,claude,defocus_blur,confidence
                2,87.03703703703704,claude,defocus_blur,confidence
                3,84.4,claude,defocus_blur,confidence
                4,82.0,claude,defocus_blur,confidence
                5,81.875,claude,defocus_blur,confidence
			};
			\addplot+[accuracy_color, no marks] table[x=level, y=metric, col sep=comma] {
				level, metric, model, etc, type
				0,79.26829268292683,claude,defocus_blur,accuracy
                1,61.53846153846154,claude,defocus_blur,accuracy
                2,59.25925925925925,claude,defocus_blur,accuracy
                3,40.0,claude,defocus_blur,accuracy
                4,36.0,claude,defocus_blur,accuracy
                5,25.0,claude,defocus_blur,accuracy
			};
		\end{axis}
	\end{tikzpicture}
    \begin{tikzpicture}
        \begin{axis}[xlabel={Severity}, ymin=0.0, ymax=100.0, width=0.48\linewidth, label style={font=\tiny}, tick label style={font=\tiny}, major tick length=\MajorTickLength, inner sep=1pt, yticklabels=\empty]
			\addplot+[confidence_color, no marks] table[x=level, y=metric, col sep=comma] {
				level, metric, model, etc
				0,89.35365853658537,claude,jpeg_compression,confidence
                1,88.07692307692308,claude,jpeg_compression,confidence
                2,89.42307692307692,claude,jpeg_compression,confidence
                3,86.2,claude,jpeg_compression,confidence
                4,84.6,claude,jpeg_compression,confidence
                5,86.88461538461539,claude,jpeg_compression,confidence
			};
			\addplot+[accuracy_color, no marks] table[x=level, y=metric, col sep=comma] {
				level, metric, model, etc, type
				0,79.26829268292683,claude,jpeg_compression,accuracy
                1,73.07692307692307,claude,jpeg_compression,accuracy
                2,61.53846153846154,claude,jpeg_compression,accuracy
                3,64.0,claude,jpeg_compression,accuracy
                4,56.00000000000001,claude,jpeg_compression,accuracy
                5,53.84615384615385,claude,jpeg_compression,accuracy
			};
		\end{axis}
	\end{tikzpicture}
    \caption{{\color{accuracy_color}Accuracy} and {\color{confidence_color}confidence} plots for the three examined models and the three corruptions for the \emph{hard} visual question answering experiment.}
    \label{fig:9plot_hard}
\end{figure}

To answer the research question, we need the models to make more and more mistakes as the severity increases to see if the confidence estimates also decrease. The models achieved fairly high accuracy scores for the easy dataset even at higher severity levels. The JUS dataset \citep{overconfidence} makes it possible to test the limits of these models by asking them nearly impossible questions. It should be emphasized that we are not interested in the actual accuracy of the models but rather in their calibration. Figure \ref{fig:9plot_hard} summarizes the accuracy and confidence scores in this task.

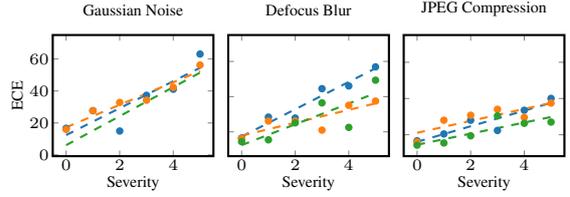
\begin{figure}[t]
    \centering
    \begin{tikzpicture}
		\begin{axis}[ylabel={ECE}, xlabel={Severity}, ymax=75, width=0.48\linewidth, label style={font=\tiny}, tick label style={font=\tiny}, major tick length=\MajorTickLength, inner sep=1pt, title={\tiny Gaussian Noise}]
			\addplot+[c0, only marks, mark size=1pt, mark options={fill=c0}] table[x=level, y=ece] {\dataHardVQAClaudeGaussianNoise};
			\addplot+[c0, no marks, dashed] table[y={create col/linear regression={x=level, y=ece}}] {\dataHardVQAClaudeGaussianNoise};
			
			\addplot+[c1, only marks, mark size=1pt, mark options={fill=c1}] table[x=level, y=ece] {\dataHardVQAGeminiGaussianNoise};
			\addplot+[c1, no marks, dashed] table[y={create col/linear regression={x=level, y=ece}}] {\dataHardVQAGeminiGaussianNoise};
			
			\addplot+[c2, only marks, mark=* mark size=1pt, mark options={fill=c1}] table[x=level, y=ece] {\dataHardVQAGPTGaussianNoise};
			\addplot+[c2, no marks, dashed] table[y={create col/linear regression={x=level, y=ece}}] {\dataHardVQAGPTGaussianNoise};
		\end{axis}
	\end{tikzpicture}
    \begin{tikzpicture}
		\begin{axis}[xlabel={Severity}, ymax=75, width=0.48\linewidth, label style={font=\tiny}, tick label style={font=\tiny}, major tick length=\MajorTickLength, inner sep=1pt, title={\tiny Defocus Blur}, yticklabels=\empty]
			\addplot+[c0, only marks, mark size=1pt, mark options={fill=c0}] table[x=level, y=ece] {\dataHardVQAClaudeDefocusBlur};
			\addplot+[c0, no marks, dashed] table[y={create col/linear regression={x=level, y=ece}}] {\dataHardVQAClaudeDefocusBlur};
			
			\addplot+[c1, only marks, mark size=1pt, mark options={fill=c1}] table[x=level, y=ece] {\dataHardVQAGeminiDefocusBlur};
			\addplot+[c1, no marks, dashed] table[y={create col/linear regression={x=level, y=ece}}] {\dataHardVQAGeminiDefocusBlur};
			
			\addplot+[c2, only marks, mark=*, mark size=1pt, mark options={fill=c2}] table[x=level, y=ece] {\dataHardVQAGPTDefocusBlur};
			\addplot+[c2, no marks, dashed] table[y={create col/linear regression={x=level, y=ece}}] {\dataHardVQAGPTDefocusBlur};
		\end{axis}
	\end{tikzpicture}
    \begin{tikzpicture}
		\begin{axis}[xlabel={Severity}, ymax=75, width=0.48\linewidth, label style={font=\tiny}, tick label style={font=\tiny}, major tick length=\MajorTickLength, inner sep=1pt, title={\tiny JPEG Compression}, yticklabels=\empty]
			\addplot+[c0, only marks, mark size=1pt, mark options={fill=c0}] table[x=level, y=ece] {\dataHardVQAClaudeJPEGCompression};
			\addplot+[c0, no marks, dashed] table[y={create col/linear regression={x=level, y=ece}}] {\dataHardVQAClaudeJPEGCompression};
			
			\addplot+[c1, only marks, mark size=1pt, mark options={fill=c1}] table[x=level, y=ece] {\dataHardVQAGeminiJPEGCompression};
			\addplot+[c1, no marks, dashed] table[y={create col/linear regression={x=level, y=ece}}] {\dataHardVQAGeminiJPEGCompression};
			
			\addplot+[c2, only marks, mark=*, mark size=1pt, mark options={fill=c2}] table[x=level, y=ece] {\dataHardVQAGPTJPEGCompression};
			\addplot+[c2, no marks, dashed] table[y={create col/linear regression={x=level, y=ece}}] {\dataHardVQAGPTJPEGCompression};
		\end{axis}
	\end{tikzpicture}
    \caption{ECE scores for {\color{c0} Claude}, {\color{c1} Gemini}, and {\color{c2} GPT-V4} for each corruption type for different severity levels in the \emph{hard} visual question answering experiment. %
    }
    \label{fig:ECE_hard}
\end{figure}

Compared to the easy task in Figure \ref{fig:9plot_easy}, we see lower accuracy scores, but more importantly, it is more visible that the gap between accuracy and confidence widens as we increase severity. This effect is apparent when we look at the ECE scores in  Figure \ref{fig:ECE_hard}.

\begin{table}[t]
\resizebox{\linewidth}{!}{%
\begin{tabular}{ l c c c }
\toprule
& {\color{c0}Claude} & {\color{c1}Gemini} & {\color{c2}GPT-V4} \\ \midrule
Gaussian Noise & 0.77 & 0.94 & 0.87 \\ 
Defocus Blur & 0.95 & 0.70 & 0.68 \\ 
JPEG Compression & 0.84 & 0.71 & 0.73 \\ \bottomrule
\end{tabular}}
\caption{R$^2$ values of ECE vs. severity trends in Fig.  \ref{fig:ECE_hard}.}
\label{tab:r_squared_hard}
\end{table}

\begin{figure}[!t]
    \centering
    \begin{tikzpicture}
		\begin{axis}[ylabel={Refusal Rate}, xlabel={Severity}, ymax=1.0, width=0.48\linewidth, label style={font=\tiny}, tick label style={font=\tiny}, major tick length=\MajorTickLength, inner sep=1pt, title={\tiny Gaussian Noise}]
			\addplot+[c2, no marks] table[x=level, y=refrate, col sep=comma] {
				refrate, level, model, etc
				0.0689655172413793,0,gpt,gaussian_noise
                0.206896551724138,1,gpt,gaussian_noise
                0.275862068965517,2,gpt,gaussian_noise
                0.413793103448276,3,gpt,gaussian_noise
                0.517241379310345,4,gpt,gaussian_noise
                0.689655172413793,5,gpt,gaussian_noise
			};
			\addplot+[c0, no marks] table[x=level, y=refrate, col sep=comma] {
				refrate, level, model, etc
				0.103448275862069,0,claude,gaussian_noise
                0.0689655172413793,1,claude,gaussian_noise
                0.137931034482759,2,claude,gaussian_noise
                0.103448275862069,3,claude,gaussian_noise
                0.172413793103448,4,claude,gaussian_noise
                0.310344827586207,5,claude,gaussian_noise			
			};
			\addplot+[c1, no marks] table[x=level, y=refrate, col sep=comma] {
				refrate, level, model, etc
				0.0,0,gemini,gaussian_noise
                0.0,1,gemini,gaussian_noise
                0.0,2,gemini,gaussian_noise
                0.0689655172413793,3,gemini,gaussian_noise
                0.0689655172413793,4,gemini,gaussian_noise
                0.172413793103448,5,gemini,gaussian_noise
			};
		\end{axis}
	\end{tikzpicture}
    \begin{tikzpicture}
		\begin{axis}[xlabel={Severity}, ymax=1.0, width=0.48\linewidth, label style={font=\tiny}, tick label style={font=\tiny}, major tick length=\MajorTickLength, inner sep=1pt, title={\tiny Defocus Blur}, yticklabels=\empty]]
			\addplot+[c2, no marks] table[x=level, y=refrate, col sep=comma] {
				refrate, level, model, etc
				0.0689655172413793,0,gpt,defocus_blur
                0.241379310344828,1,gpt,defocus_blur
                0.448275862068966,2,gpt,defocus_blur
                0.517241379310345,3,gpt,defocus_blur
                0.517241379310345,4,gpt,defocus_blur
                0.517241379310345,5,gpt,defocus_blur				
			};
			\addplot+[c0, no marks] table[x=level, y=refrate, col sep=comma] {
				refrate, level, model, etc
				0.103448275862069,0,claude,defocus_blur
                0.103448275862069,1,claude,defocus_blur
                0.0689655172413793,2,claude,defocus_blur
                0.137931034482759,3,claude,defocus_blur
                0.137931034482759,4,claude,defocus_blur
                0.172413793103448,5,claude,defocus_blur
			};
			\addplot+[c1, no marks] table[x=level, y=refrate, col sep=comma] {
				refrate, level, model, etc
				0.0,0,gemini,defocus_blur
                0.0344827586206897,1,gemini,defocus_blur
                0.0344827586206897,2,gemini,defocus_blur
                0.0,3,gemini,defocus_blur
                0.0344827586206897,4,gemini,defocus_blur
                0.0689655172413793,5,gemini,defocus_blur
			};	
		\end{axis}
	\end{tikzpicture}
    \begin{tikzpicture}
		\begin{axis}[xlabel={Severity}, ymax=1.0, width=0.48\linewidth, label style={font=\tiny}, tick label style={font=\tiny}, major tick length=\MajorTickLength, inner sep=1pt, title={\tiny JPEG Compression}, yticklabels=\empty]]
			\addplot+[c2, no marks] table[x=level, y=refrate, col sep=comma] {
				refrate, level, model, etc
				0.0689655172413793,0,gpt,jpeg_compresion
                0.0689655172413793,1,gpt,jpeg_compresion
                0.0689655172413793,2,gpt,jpeg_compresion
                0.206896551724138,3,gpt,jpeg_compresion
                0.172413793103448,4,gpt,jpeg_compresion
                0.172413793103448,5,gpt,jpeg_compresion			
			};
			\addplot+[c0, no marks] table[x=level, y=refrate, col sep=comma] {
				refrate, level, model, etc
				0.103448275862069,0,claude,jpeg_compresion
                0.103448275862069,1,claude,jpeg_compresion
                0.103448275862069,2,claude,jpeg_compresion
                0.137931034482759,3,claude,jpeg_compresion
                0.137931034482759,4,claude,jpeg_compresion
                0.103448275862069,5,claude,jpeg_compresion
			};
			\addplot+[c1, no marks] table[x=level, y=refrate, col sep=comma] {
				refrate, level, model, etc
				0.0,0,gemini,jpeg_compresion
                0.0344827586206897,1,gemini,jpeg_compresion
                0.0,2,gemini,jpeg_compresion
                0.0344827586206897,3,gemini,jpeg_compresion
                0.0,4,gemini,jpeg_compresion
                0.0344827586206897,5,gemini,jpeg_compresion
			};
		\end{axis}
	\end{tikzpicture}
    \caption{Refusal rates for {\color{c0} Claude}, {\color{c1} Gemini}, and \mbox{{\color{c2} GPT-V4}} across different severity levels and corruptions in the \emph{hard} visual question answering experiment.}
    \label{fig:refusals_hard}
\end{figure}

The most visible difference between Figures \ref{fig:ECE_easy} and \ref{fig:ECE_hard} is for JPEG compression and defocus blur corruptions. The models become more miscalibrated at higher severity levels. There is not as much difference between the models for Gaussian noise as in the easy dataset. We can also see that GPT-4V still achieves the lowest ECE scores, but the models show a more similar behaviour.

Table \ref{tab:r_squared_hard} shows the R$^2$ values for the linear regression lines in Figure \ref{fig:ECE_hard}. All of the values are around or above 0.7 indicating that the regression model explains the data well. This strengthens the visual intuition that we got from comparing Figures \ref{fig:9plot_easy} and \ref{fig:9plot_hard}. We have stronger evidence that the models become more and more miscalibrated as we increase the severity of the corruption. The calibration plots shown in Appendix \ref{ap:calibration_plots} show that GPT-4V again performed much better than the other two models.

As with the easy dataset, we can also examine the refusal rates in Figure \ref{fig:refusals_hard}. For Gemini and Claude, they are around the same as in the easier dataset but for GPT-4V, they are much higher. GPT-4V performed best on the hard dataset both in terms of ECE and the calibration plots, so we see that refusing to answer a question instead of making a wrong guess can improve a model's calibration.

\subsection{Counting Task} \label{subsec:counting_task}

\begin{figure}[t]
    \centering
    \begin{tikzpicture}
		\begin{axis}[ylabel={Coverage (\%)}, xlabel={Severity}, ymax=100, width=0.48\linewidth, label style={font=\tiny}, tick label style={font=\tiny}, major tick length=\MajorTickLength, xtick pos=bottom, inner sep=1pt, title={\tiny Gaussian Noise}]
			\addplot+[c0, only marks, mark size=1pt, mark options={fill=c0}] table[x=level, y=acc] {\dataCountingClaudeGaussianNoise};
			\addplot+[c0, no marks, dashed] table[y={create col/linear regression={x=level, y=acc}}] {\dataCountingClaudeGaussianNoise};
			
			\addplot+[c1, only marks,  mark size=1pt, mark options={fill=c1}] table[x=level, y=acc] {\dataCountingGeminiGaussianNoise};
			\addplot+[c1, no marks, dashed] table[y={create col/linear regression={x=level, y=acc}}] {\dataCountingGeminiGaussianNoise};
			
			\addplot+[c2, only marks, mark=*, mark size=1pt, mark options={fill=c2}] table[x=level, y=acc] {\dataCountingGPTGaussianNoise};
			\addplot+[c2, no marks, dashed] table[y={create col/linear regression={x=level, y=acc}}] {\dataCountingGPTGaussianNoise};

            \addplot+[gray, no marks, dotted] coordinates {(0, 95) (5,95)};
		\end{axis}
	\end{tikzpicture}
    \begin{tikzpicture}
		\begin{axis}[xlabel={Coverage}, ymax=100, width=0.48\linewidth, label style={font=\tiny}, tick label style={font=\tiny}, major tick length=\MajorTickLength, xtick pos=bottom, inner sep=1pt, title={\tiny Defocus Blur}, yticklabels=\empty]
			\addplot+[c0, only marks, mark size=1pt , mark options={fill=c0}] table[x=level, y=acc] {\dataCountingClaudeDefocusBlur};
			\addplot+[c0, no marks, dashed] table[y={create col/linear regression={x=level, y=acc}}] {\dataCountingClaudeDefocusBlur};
			
			\addplot+[c1, only marks, mark size=1pt, mark options={fill=c1}] table[x=level, y=acc] {\dataCountingGeminiDefocusBlur};
			\addplot+[c1, no marks, dashed] table[y={create col/linear regression={x=level, y=acc}}] {\dataCountingGeminiDefocusBlur};
			
			\addplot+[c2, only marks, mark=*, mark size=1pt, mark options={fill=c2}] table[x=level, y=acc] {\dataCountingGPTDefocusBlur};
			\addplot+[c2, no marks, dashed] table[y={create col/linear regression={x=level, y=acc}}] {\dataCountingGPTDefocusBlur};

            \addplot+[gray, no marks, dotted] coordinates {(0, 95) (5,95)};
		\end{axis}
	\end{tikzpicture}
    \begin{tikzpicture}
		\begin{axis}[xlabel={Coverage}, ymax=100, width=0.48\linewidth, label style={font=\tiny}, tick label style={font=\tiny}, major tick length=\MajorTickLength, xtick pos=bottom, inner sep=1pt, title={\tiny JPEG Compression}, yticklabels=\empty]
			\addplot+[c0, only marks, mark size=1pt, mark options={fill=c0}] table[x=level, y=acc] {\dataCountingClaudeJPEGCompression};
			\addplot+[c0, no marks, dashed] table[y={create col/linear regression={x=level, y=acc}}] {\dataCountingClaudeJPEGCompression};
			
			\addplot+[c1, only marks, mark size=1pt, mark options={fill=c1}] table[x=level, y=acc] {\dataCountingGeminiJPEGCompression};
			\addplot+[c1, no marks, dashed] table[y={create col/linear regression={x=level, y=acc}}] {\dataCountingGeminiJPEGCompression};
			
			\addplot+[c2, only marks, mark=*, mark size=1pt, mark options={fill=c2}] table[x=level, y=acc] {\dataCountingGPTJPEGCompression};
			\addplot+[c2, no marks, dashed] table[y={create col/linear regression={x=level, y=acc}}] {\dataCountingGPTJPEGCompression};

            \addplot+[gray, no marks, dotted] coordinates {(0, 95) (5,95)};
		\end{axis}
	\end{tikzpicture}
    \caption{Coverage (confidence interval accuracy) scores for the counting experiment for {\color{c0} Claude}, {\color{c1} Gemini}, and {\color{c2} GPT-V4}. The dotted line at the top represents the 95\% accuracy which would be expected for a perfectly calibrated model.}
    \label{fig:count_acuracy}
\end{figure}
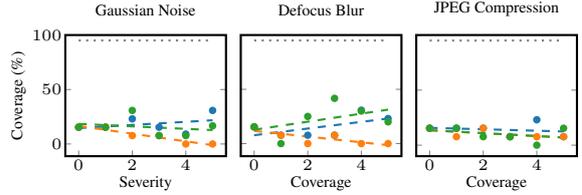

The JUS dataset contains hard counting tasks that were evaluated using a different prompt described in Appendix \ref{ap:appendix}. In this task, the model was asked to output a 95\% confidence interval. The answer was recorded as correct if that interval contained the actual prediction. There was one picture illustrated in Appendix \ref{ap:appendixB} where there was no correct answer. It is impossible to count the bamboo trees without seeing their trunks as many of the visible branches could belong to the same bamboo tree. The results of this experiment are shown in Figure \ref{fig:count_acuracy}.

For a perfectly calibrated model, we would expect that a 95\% confidence interval is correct 95\% of the time. We can see that the models perform below 25\% accuracy most of the time. There is not as much consistency in the linear regression lines as in the previous two tasks, most likely due to the models' poor performance on all severity levels and the high variance from the low number of test images.

We can look at Table \ref{tab:r_squared_count} containing the R$^2$ values for the lines but we get much lower values than in the previous two tasks. The models are unable to answer the questions even on the original dataset, so increasing the severity of the corruption does not have an effect.

It is important to note that the models seldom refused to provide a response during this task. Out of the 208 times each model was queried (13 questions, 3 corruptions, 5 severity levels plus the original dataset), GPT-4V refused to answer 9 times, Gemini 2 times, and Claude 0 times. The low accuracies show that the models responded even when the 95\% confidence interval was purely guessed.

\begin{table}[t]
\resizebox{\linewidth}{!}{%
\begin{tabular}{ l c c c }
\toprule
& {\color{c0}Claude} & {\color{c1}Gemini} & {\color{c2}GPT-4V} \\ \midrule
Gaussian Noise & 0.13 & 0.61 & 0.22 \\ 
Defocus Blur & 0.35 & 0.47 & 0.13 \\ 
JPEG Compression & 0.31 & 0.02 & 0.31 \\ \bottomrule
\end{tabular}}
\caption{R$^2$ values of accuracy vs. severity trends in Fig. \ref{fig:count_acuracy}.}
\label{tab:r_squared_count}
\end{table}

Interestingly, the models often had an exact guess that was reasonably close to the right answer, but their confidence interval was so small that it almost never contained the true value. This shows the models' good visual capabilities even on hard images, but also signals their bad calibration as they were not capable of formulating an accurate 95\% confidence interval based on a close estimate.

\section{Discussion}\label{sec:discussion}
Overall we found that increased corruption severity had a negative impact on the three examined models' accuracy and calibration. When the corruption level gradually became higher and the models' accuracy started to decrease, it was not accompanied by decreasing confidence scores. We also found that models that refuse to answer at a higher rate can achieve better accuracy and ECE scores. Our other main finding is that models are generally overconfident in their responses and output high confidence scores in most of their responses. This overconfidence was present in all three experiments but it was the most severe in the counting problems.

\textbf{Interpretation of Results}. Our findings about overconfident models and their high confidence scores are in line with \citet{overconfidence} and \citet{verbalized_uncertinty}. Both of them found that the majority of the confidence scores of LLMs and VLMs fall within the [80, 100] range. While \citet{verbalized_uncertinty} looked at the performance of LLMs in different reasoning tasks (commonsense, arithmetic, symbolic), \citet{overconfidence} also examined the performance of VLMs in visual question answering tasks. Our results show that this characteristic of VLMs persists when they are tested on corrupted images. Higher corruption levels worsening calibration was also found by \citet{hendrycks2019benchmarking} who tested different neural network architectures designed for image classification. We show that their findings can be extended to the realm of VLMs.

As mentioned before, one possible explanation for this overconfidence is the RLHF fine-tuning of these models. It rewards answers that sound more confident so the model learns to express its responses using confident language which influences the confidence scores of verbalized uncertainty.

Examining the the number of cases where the models refused to answer, we found that higher refusal rates can help the model's calibration as it is nearly impossible to give a correct answer to some highly distorted images. Especially with Gaussian noise, there were times when the models were fooled by the noise and output completely unrelated answers to the images. This happened less with JPEG compression as it was a less severe corruption than the other two. In case of defocus blur, the models were more likely to recognise the heavy blurring effect on the image and refuse to respond to the question. Images corrupted with Gaussian noise were less likely to be recognised as corrupted and in some cases they were even confused with pointillistic paintings which is a painting technique from the late 19$^{\text{th}}$ century using small colourful dots that form an image when viewed from a distance. For some examples of model responses, we refer to Appendix \ref{ap:appendixE}.

\textbf{Future Research}. Apart from increasing the number of images in the dataset, there are other things that could be explored in the topic of uncertainty estimation in VLMs. Different prompting strategies, such as chain-of-thought reasoning or top-k explored by \citet{verbalized_uncertinty} could yield different results. These can be altered so the models are more restricted in their answers making automated data gathering easier. Apart from verbalized uncertainty, there exist other, sampling-based techniques for uncertainty estimation \citep{tian2023just} that could be applied to VLMs. 

The overconfidence of RLHF-based LLMs is present in multiple studies \citep{overconfidence, verbalized_uncertinty} but it would be interesting to explore if this overconfidence in VLMs could be treated with temperature scaling in the same way as in \citet{rlhftempscale}. The APIs provided for the three investigated VLMs offer the ability to manipulate the model's temperature.

\citet{corruptions_repo} defines 15 corruption types, but we only tested three. Studying the effect of the others could reveal more differences between the models and their robustness to different corruptions.

\section{Conclusions and Future Work}\label{sec:conclusion}

The key conclusions that we obtained from this work are the following:

\begin{itemize}
    \item VLMs are overconfident.
    They often express their confidence in the range of [80, 100] even when this is not reflected in their accuracy.
    
    \item Increased corruption severity increases the ECE. 
    When the models started making mistakes due to the increasingly corrupted images, their confidence did not decrease at the same pace which caused the ECE to go up. This is the main finding of the paper, as it answers our research question.

    \item There are differences in the calibration of state-of-the-art VLMs and the models are more robust to some corruptions than others.
    GPT-4V outperformed the other two models in the visual question-answering experiments, and JPEG compression was better handled by all of the models than Gaussian noise and defocus blur.

    \item Higher refusal rates can improve calibration.
    We see that when the model recognises that we are asking an impossible question and refuses to answer, it prevents itself from providing hallucinated answers and improves its calibration. GPT-4V also performed better in this regard than the other two models.

    \item VLMs were especially miscalibrated when they were asked to express their answer in a 95\% confidence interval.
    Their accuracy in the counting experiment did not even come close to 95\%, even when their initial guess for the exact number of objects was quite close to the answer.

\end{itemize}

From these results, we can see that there are many things that can be improved when it comes to the calibration of VLMs. In the current state of things, users are often presented with confident wrong answers which undermine the trust in these models. This paper contributes to the research in uncertainty estimation of VLMs and points out the shortcomings of these models with respect to their calibration. Better-calibrated models would be beneficial to millions of users as these models are already widely used by the general public.

\section{Limitations}
As the models output high confidence scores, the lower confidence bins were underrepresented in the calibration plots. This issue could be solved with more images in the datasets, but there were some limitations on the number of images that the models could be tested on. For the easy visual question answering experiment, there were tens of thousands more images available from the dataset by \citet{VQA} and \citet{balanced_vqa_v2}. However, all of the images were used from the JUS dataset, which put a limit on the number of images in the hard VQA experiment and the counting experiment.

Since a correct answer to a question could be phrased in multiple ways, the answers had to be manually checked, which made the data-gathering process time-consuming. The used APIs also had a limit on the number of requests per minute, which prevented large-scale testing.

One way to automate the check for the correctness of the answers could be to use an LLM to check the semantic equivalence of the correct reference answer and the response provided by the model. This was not a suitable approach in this paper, as these methods still have limited correlation (Spearman rank correlation $\rho < 0.7$) with manual assessment \cite{manas2024improving}. Another way could be to use better prompts that restrict the model to one or two-word answers that are easier to check automatically but that would put a limit on the complexity of the tested questions.

Lastly, we found that some models are more likely to refuse to answer for images with severe corruptions. These samples can therefore not be included in the answer, which can decrease error rate and decrease the recorded miscalibration. This may be considered acceptable, as not giving an answer may be preferable over a random guess, but it is not as good as giving a correct answer. There are methods that attempt to minimize the refusal rate and get more correct answers \cite{srinivasan2024selective}, but those were not considered in this study. 

\section{Ethical Considerations}

In this paper we show that VLMs verbalized uncertainty is prone to severe degradation under input image corruption, which adds to other common problems with VLMs like incorrect predictions and hallucinations, this raises ethical concerns on their use, as these models are effectively not able to identify when they do not know or cannot answer a prompt.

Users of these models are recommended to always double check with a human any kind of output that is given by VLM, as they are not trustworthy, and when used for critical applications, humans can be hurt.

Our work shows that GPT4V is able to detect Gaussian noise corruptions, which indicates that its developers might have included this in its training set, but not other kinds of corruptions, showcasing the limitations of VLMs as only capabilities considered during training are available during inference/deployment. More research is needed to further detect other kinds of input image corruptions.

\bibliography{thesis_original_format/literature_bachelorthesis}
\newpage
\onecolumn
\appendix

\section{Calibration Plots}\label{ap:calibration_plots}

Since one metric like the ECE can hide the nuances in the model's behaviour, we can make a calibration plot for each corruption. We calculate the model's average confidence in a confidence bin and plot its accuracy along the y-axis.

\begin{figure*}[h]
    \centering
    \includegraphics[width=\linewidth]{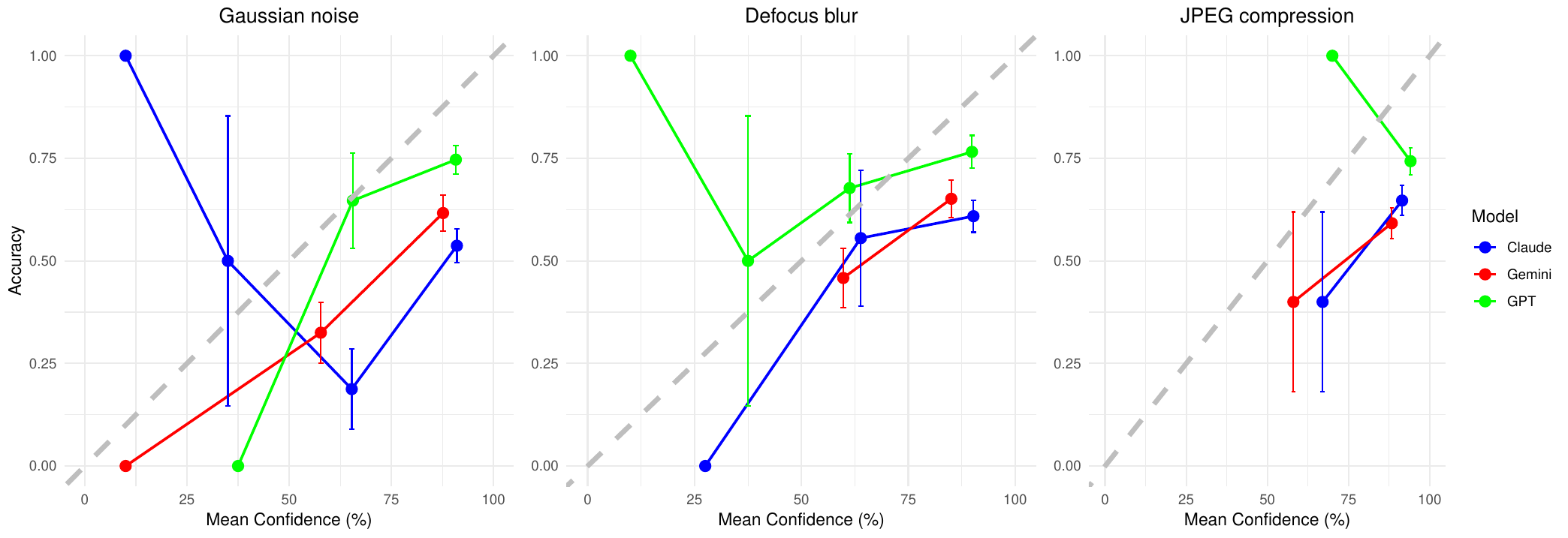}
    \caption{Calibration plots for the three examined corruption types and three models in the easy visual question answering experiment. The error bars represent the standard error and the dashed line indicates perfect calibration.}
    \label{fig:claibration_easy}
\end{figure*}

Figure \ref{fig:claibration_easy} illustrates the calibration of the three models in each corruption type for the "easy" visual question answering. Since the calibration plots for one specific severity level and corruption type are too noisy due to the low number of data points and the uneven distribution of the confidence scores, we plotted the calibration using all severity levels of a given corruption. Even with this adjustment, most of the bins contained one or two data points so we decided to use four equal bins covering the [0, 100] interval.

GPT-4V outperformed the other two models in all types of corruption. It is especially close to the dashed line indicating perfect calibration, in the defocus blur corruption. In the confidence bins where there were enough data points, indicated by the small error bars, the models show overconfidence as those points lie below the dashed line.

There are points below 50\% confidence where the error bar is very large or zero. These points occur because the models tend to output high confidence scores so lower bins have few or no data points. If there are only one or two correct answers in a confidence bin and no other data points, then the accuracy for that bin will be 100\% and the standard error will be undefined or zero since it is calculated as the standard deviation divided by the square root of the number of samples.

\begin{figure*}[h]
    \centering
    \includegraphics[width=\linewidth]{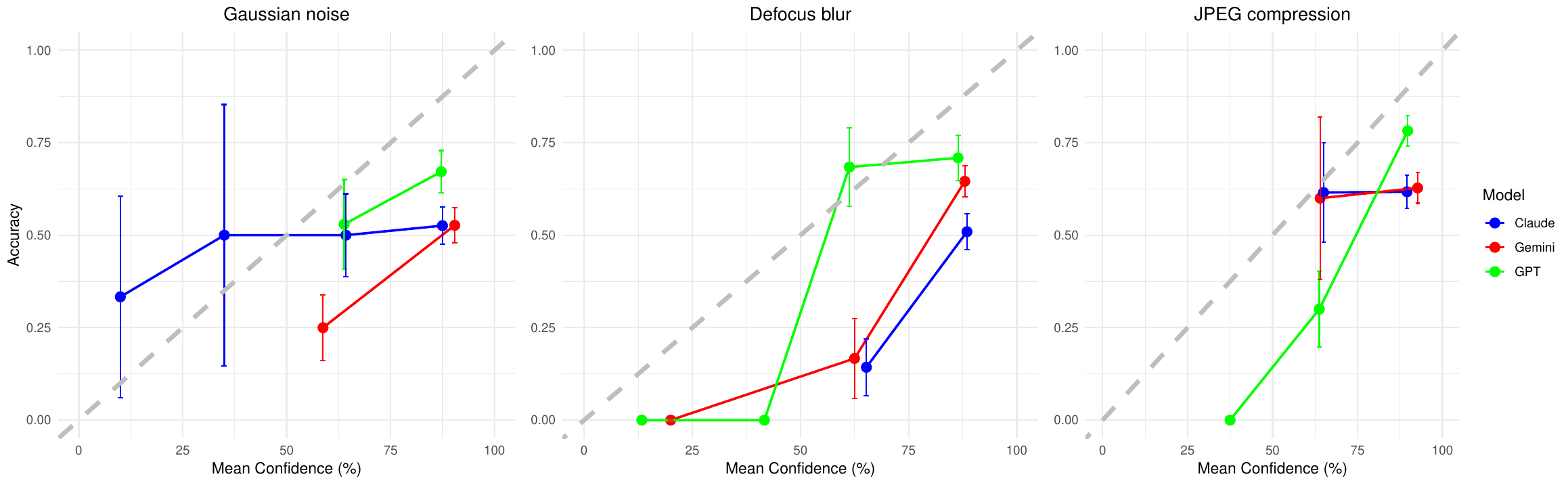}
    \caption{Calibration plots for the three examined corruption types and three models in the hard visual question answering experiment with the error bars calculated using the standard error.}
    \label{fig:claibration_hard}
\end{figure*}

Figure \ref{fig:claibration_hard} illustrates the calibration for the "hard" visual question answering. There we can again see that GPT-4V performed much better than the other two models. It should be noted that the problem of confidence bins with low or zero number of data points is still present, so the same bin size had to be used.

\section{Image data for the easy VQA and model questions}\label{ap:appendix}
To reduce the size of the Easy VQA dataset to be able to manually assess the responses we randomly selected a subset of the images and their questions. The selected images can be found in Table \ref{tab:images_easy_vqa}.

To elicit verbalized confidence we expanded the questions used. We appended either:

'\textit{Moreover, please rate your confidence in your answer between 0 and 100\%. The answer should be in the format: "Answer (confidence\%)".}' or 

'\textit{Provide your actual prediction. Moreover, please express your estimate as a 95\% confidence interval. This means you should provide a range within which you are 95\% confident the true value lies. Format your answer as: "[Lower Bound, Upper Bound]", where the lower bound is the start of the range and the upper bound is the end of the range. Ensure that this interval reflects a 95\% confidence level based on your estimation.}', depending on whether the question was a counting task. 
\begin{table}[h]
    \centering
        \caption{Randomly selected images and their corresponding questions for "easy" VQA.}
    \begin{tabular}{l l}\toprule
        Image ID & Question \\\midrule 
        COCO\_test2015\_000000341181&     Is the kitchen well lit? \\
        COCO\_test2015\_000000244073&     What color is the plane? \\
        COCO\_test2015\_000000415036&     Is there a plug near the bed? \\
        COCO\_test2015\_000000551714&     Is the person wearing gloves? \\
        COCO\_test2015\_000000084296&     How many chairs are around the table? \\
        COCO\_test2015\_000000512556&     What is the bench made of? \\
        COCO\_test2015\_000000358972&     What type of boat is that? \\
        COCO\_test2015\_000000154340&     Are they preparing food? \\
        COCO\_test2015\_000000473114&     Is this computer equipment? \\
        COCO\_test2015\_000000444844&     Is this apartment completely empty? \\
        COCO\_test2015\_000000459379&     What is the boy doing? \\
        COCO\_test2015\_000000003004&     What is the animal doing? \\
        COCO\_test2015\_000000515370&     What shape is the kite? \\
        COCO\_test2015\_000000066725&     What appliance is pictured? \\
        COCO\_test2015\_000000121284&     What time is it? \\
        COCO\_test2015\_000000420197&     What is beside the dog? \\
        COCO\_test2015\_000000471029&     Is the man right-handed? \\
        COCO\_test2015\_000000112870&     What is on the shelf above the toilet? \\
        COCO\_test2015\_000000343994&     What is in the baby's mouth? \\
        COCO\_test2015\_000000351008&     Is the horse running down the street? \\
        COCO\_test2015\_000000334624&     Is the man listening to something on his smartphone? \\
        COCO\_test2015\_000000177197&     What color is the batter wearing? \\
        COCO\_test2015\_000000407045&     Is the bench brown? \\
        COCO\_test2015\_000000110643&     How many birds? \\
        COCO\_test2015\_000000517475&     Are they using foil paper? \\
        COCO\_test2015\_000000268054&     Is the oven on? \\
        COCO\_test2015\_000000068573&     Is this indoors? \\
        COCO\_test2015\_000000262294&     What type of sweater is the man wearing? \\
        COCO\_test2015\_000000206488&     Is there money on the table? \\
        COCO\_test2015\_000000066282&     What type of place is this? \\
        COCO\_test2015\_000000166735&     Are there people in the boat? \\
        COCO\_test2015\_000000068702&     Are there picture frames in this picture? \\
        COCO\_test2015\_000000475609&     Has the ball been thrown? \\
        COCO\_test2015\_000000434294&     Is the boy wearing shoes? \\
        COCO\_test2015\_000000373079&    Is the person wearing a wedding band? \\
        COCO\_test2015\_000000022109&     What is on the sheep?\\ 
    \bottomrule
    \end{tabular}
    \label{tab:images_easy_vqa}
\end{table}    
\section{Demonstration of corruptions and an impossible question}\label{ap:appendixB}

\begin{figure}[htbp]
    \centering
    \resizebox{\linewidth}{!}{%
    \begin{tabular}{>{\arraybackslash}m{1.8cm} c c c c c c}
        \vspace{-1cm}
        Gaussian Noise:\vfill &
        \includegraphics[width=2cm]{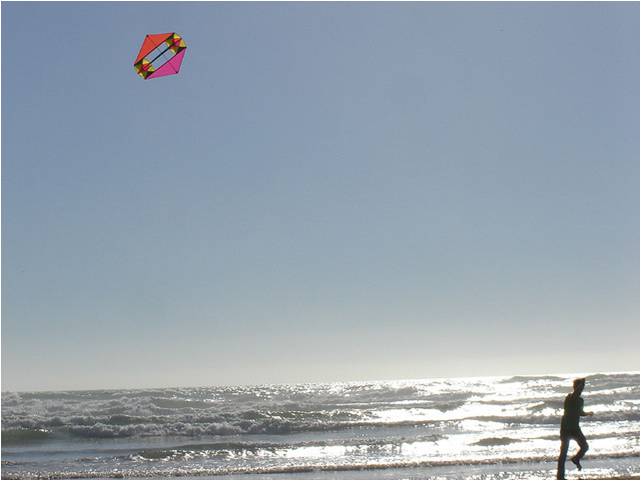} &
        \includegraphics[width=2cm]{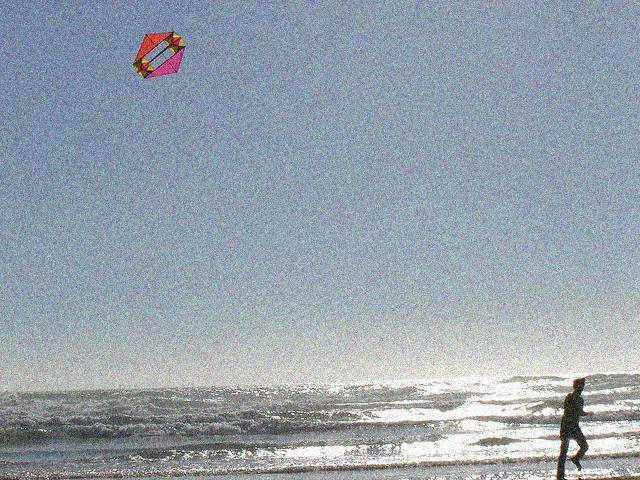} &
        \includegraphics[width=2cm]{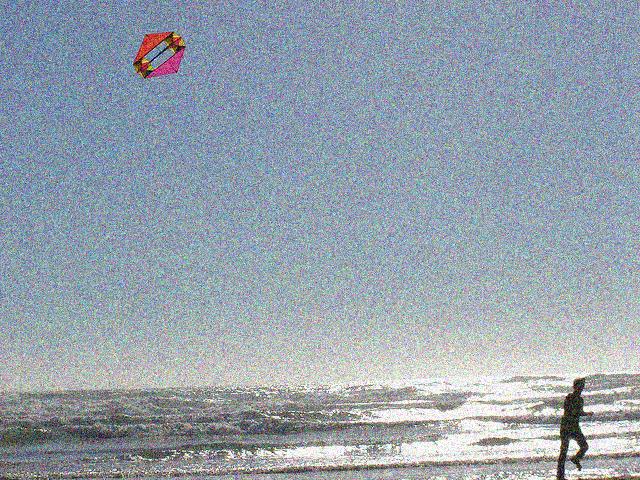} &
        \includegraphics[width=2cm]{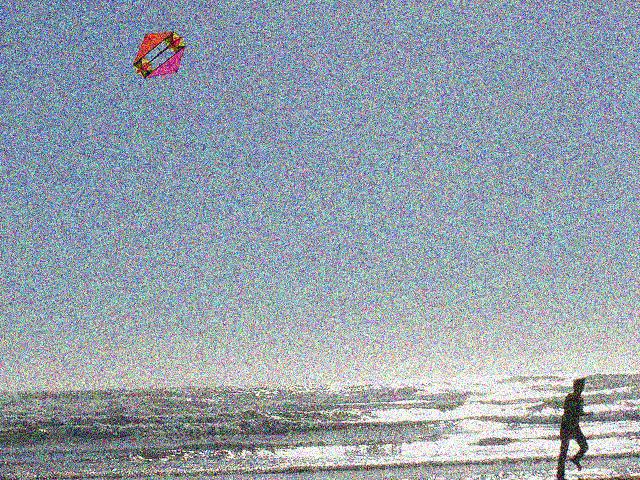} &
        \includegraphics[width=2cm]{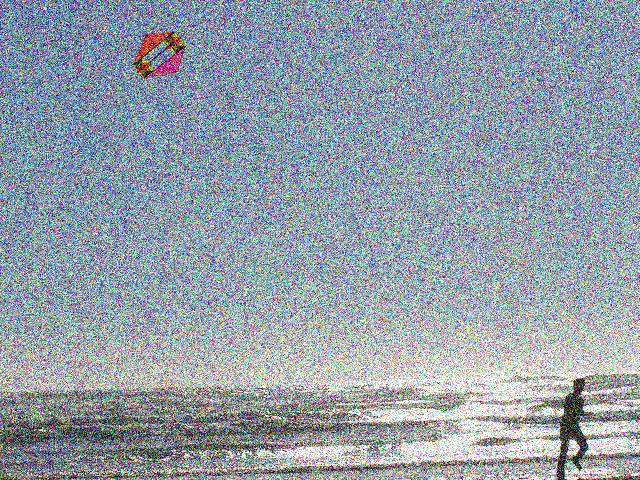} &
        \includegraphics[width=2cm]{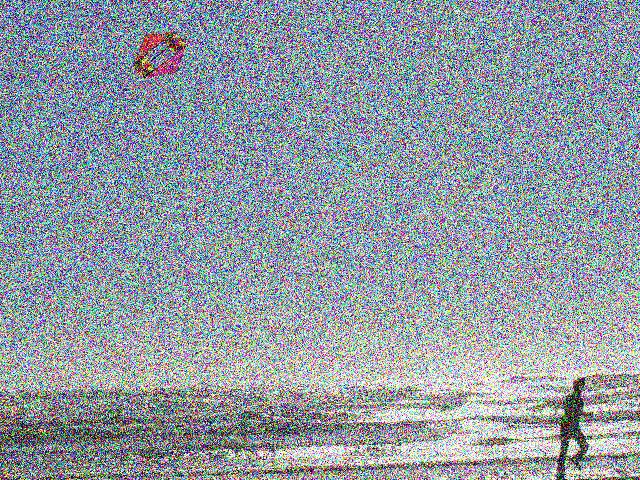} \\

        \vspace{-1cm}
        JPEG \mbox{Compression}:\vfill & 
        \includegraphics[width=2cm]{img/original_image.jpg} &
        \includegraphics[width=2cm]{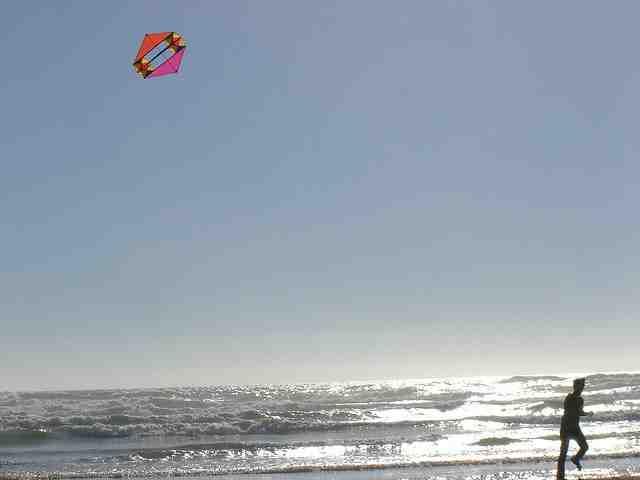} &
        \includegraphics[width=2cm]{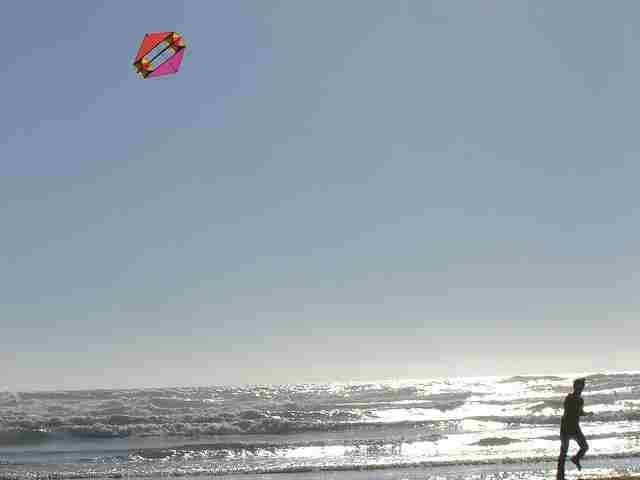} &
        \includegraphics[width=2cm]{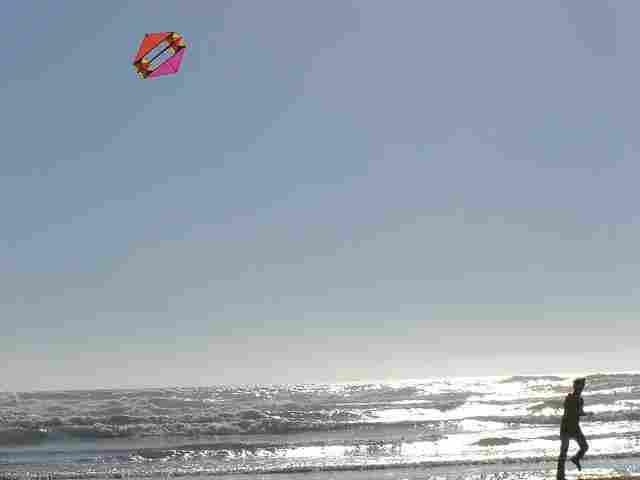} &
        \includegraphics[width=2cm]{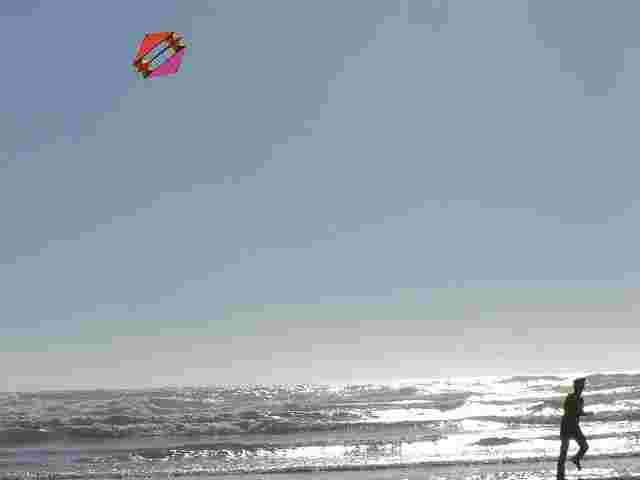} &
        \includegraphics[width=2cm]{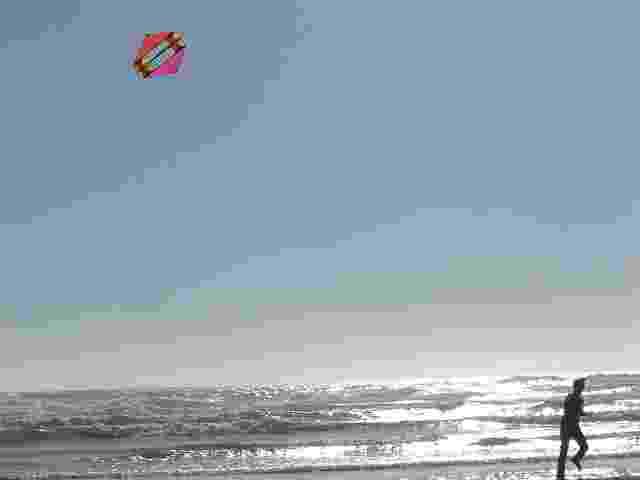} \\

        \vspace{-1cm}
        Defocus Blur:\vfill & 
        \includegraphics[width=2cm]{img/original_image.jpg} &
        \includegraphics[width=2cm]{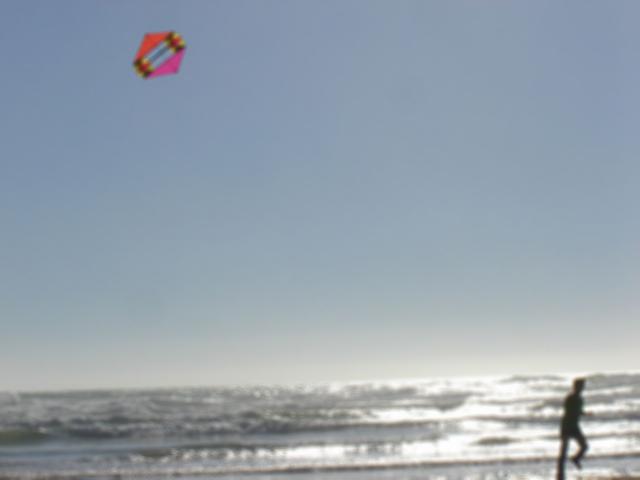} &
        \includegraphics[width=2cm]{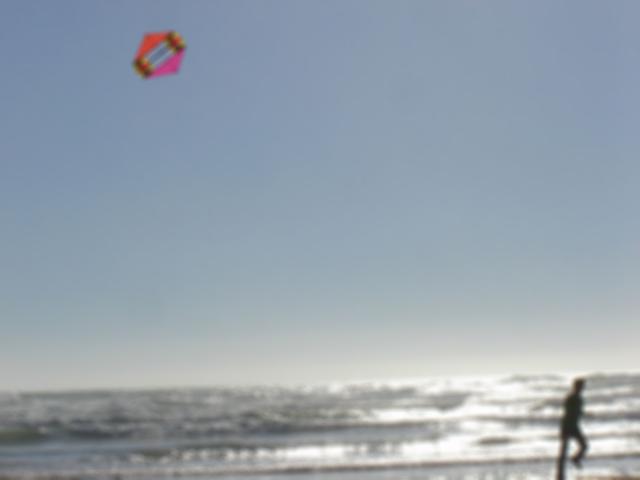} &
        \includegraphics[width=2cm]{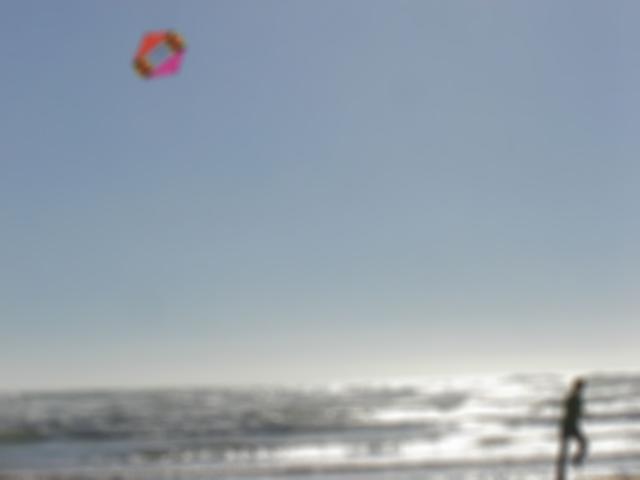} &
        \includegraphics[width=2cm]{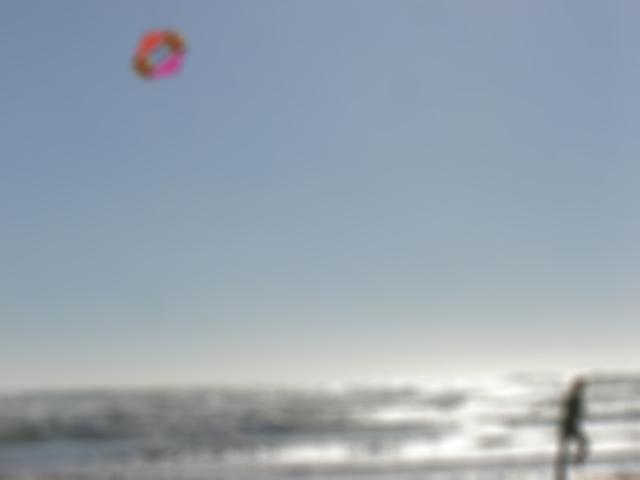} &
        \includegraphics[width=2cm]{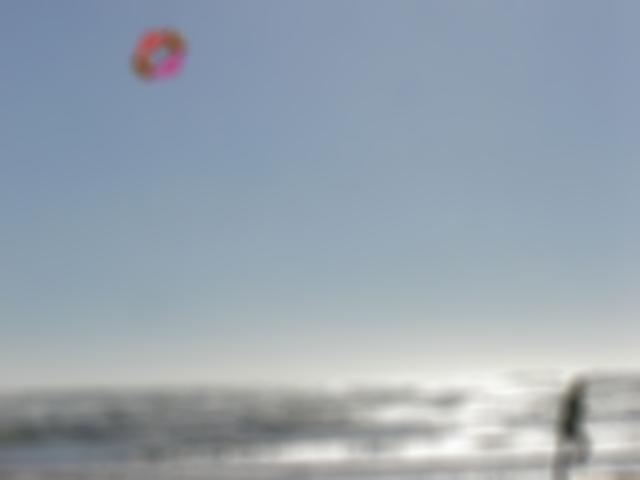} \\
        
        & Severity 0 & Severity 1 & Severity 2 & Severity 3 & Severity 4 & Severity 5 \\
    \end{tabular}}
    \caption{Demonstration of different severity levels for the three tested corruptions}
    \label{fig:overall}
\end{figure}

\begin{figure*}[h]
    \centering
        \includegraphics[width=7.7cm]{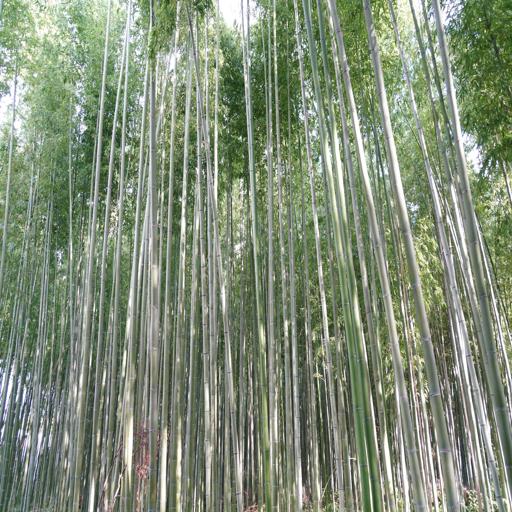}
    \caption{Impossible question: How many bamboo trees are shown in this photo?}
    \label{fig:bamboo_trees}
\end{figure*}

Figure \ref{fig:overall} illustrates the effect of the used corruptions at different severity levels. In Figure \ref{fig:bamboo_trees}, we give an example of a question that even a human should not be able to answer. As mentioned in subsection \ref{subsec:counting_task}, we expect the model not to answer. It should be noted that some of the images are so distorted at higher severity levels that they also become impossible to answer.

\newpage
\onecolumn

\section{Frequency of correct and incorrect answers in each confidence bin}\label{ap:appendixC}

\begin{figure}[h]
    \centering
    \includegraphics[width=0.7\linewidth]{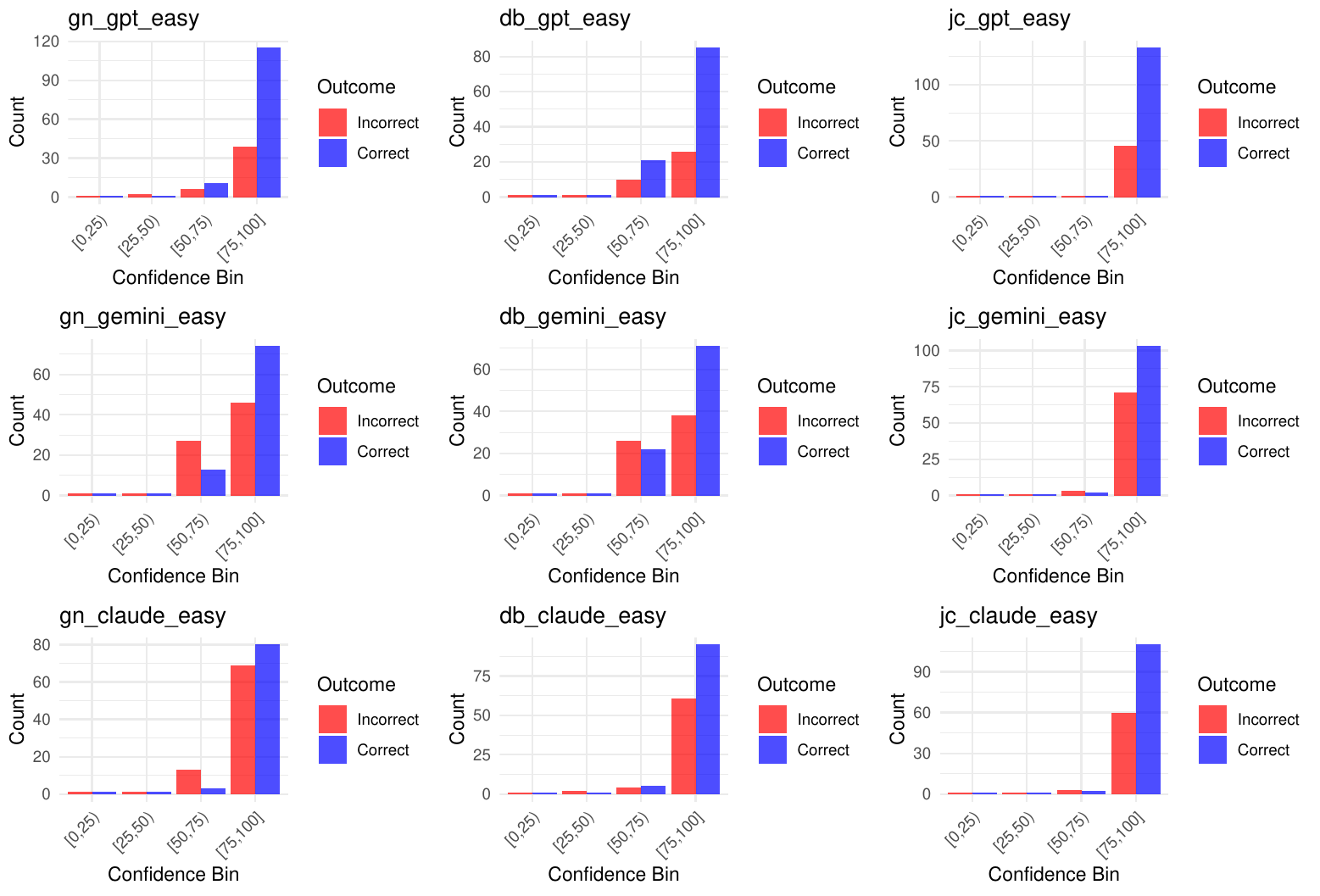}
    \caption{Histograms of confidence scores in the easy visual question answering experiment with the red and blue bars indicating the proportion of correct and incorrect responses}
    \label{fig:hist_easy}
\end{figure}

\begin{figure}[h]
    \centering
    \includegraphics[width=0.7\linewidth]{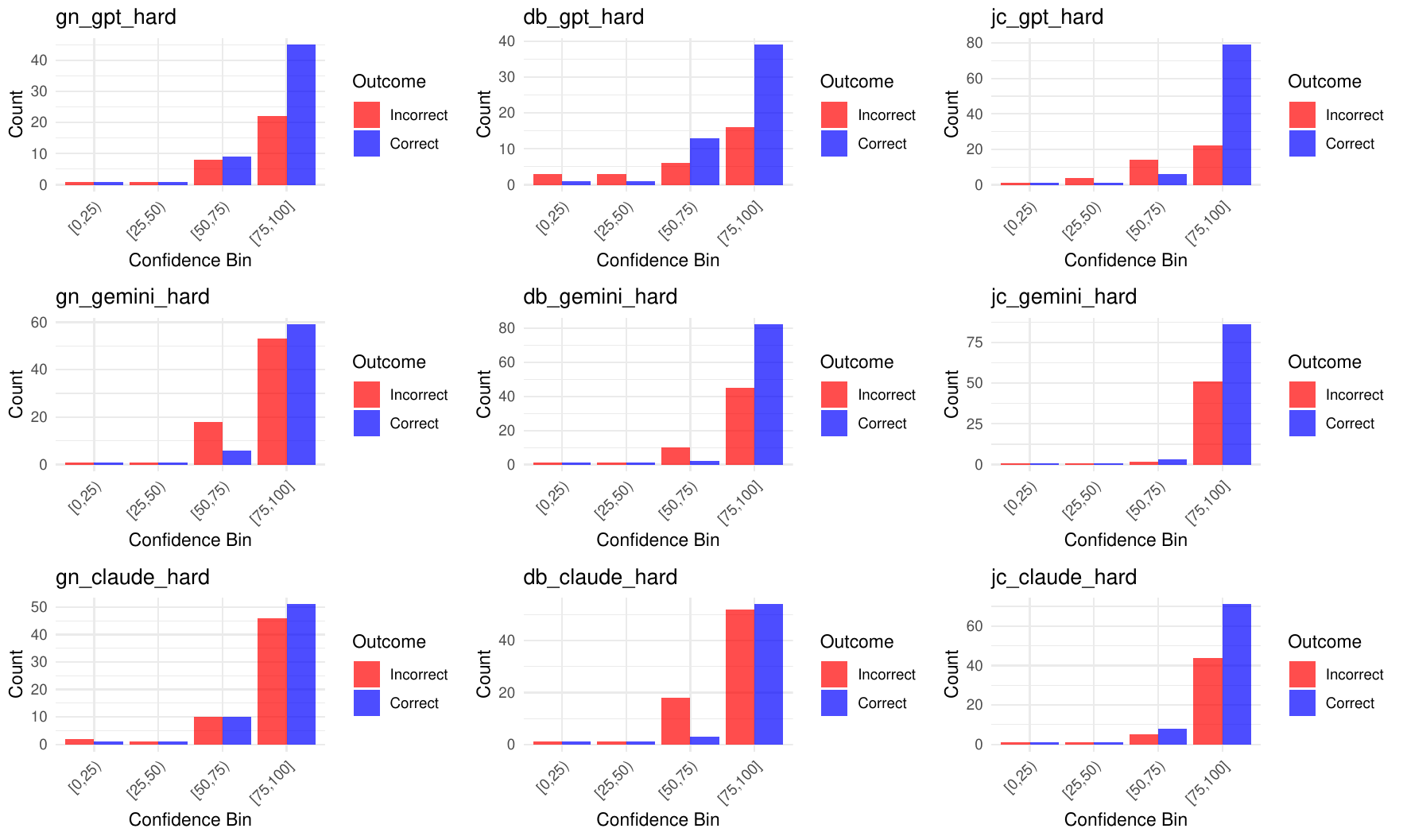}
    \caption{Histograms of confidence scores in the hard visual question answering experiment with the red and blue bars indicating the proportion of correct and incorrect responses}
    \label{fig:hist_hard}
\end{figure}

In Figures \ref{fig:hist_easy} and \ref{fig:hist_hard} 'gn' stands for Gaussian noise, 'db' stands for defocus blur and 'jc' stands for JPEG compression. There are two main things that we can observe from the two figures which influence the calibration plots. The proportion of correct and incorrect answers does not reflect the confidence bin and most of the responses fall into higher confidence bins.

\newpage
\onecolumn

\section{Frequency of all answers in each confidence bin}\label{ap:appendixD}

In the 6 Figures below, we see the frequency of confidence scores for each model in the two VQA tasks. While models give more low-confidence answers in the hard VQA task, the overwhelming majority of answers fall into the higher confidence bins.

\begin{figure}[h]
    \centering
    \includegraphics[width=0.75\linewidth]{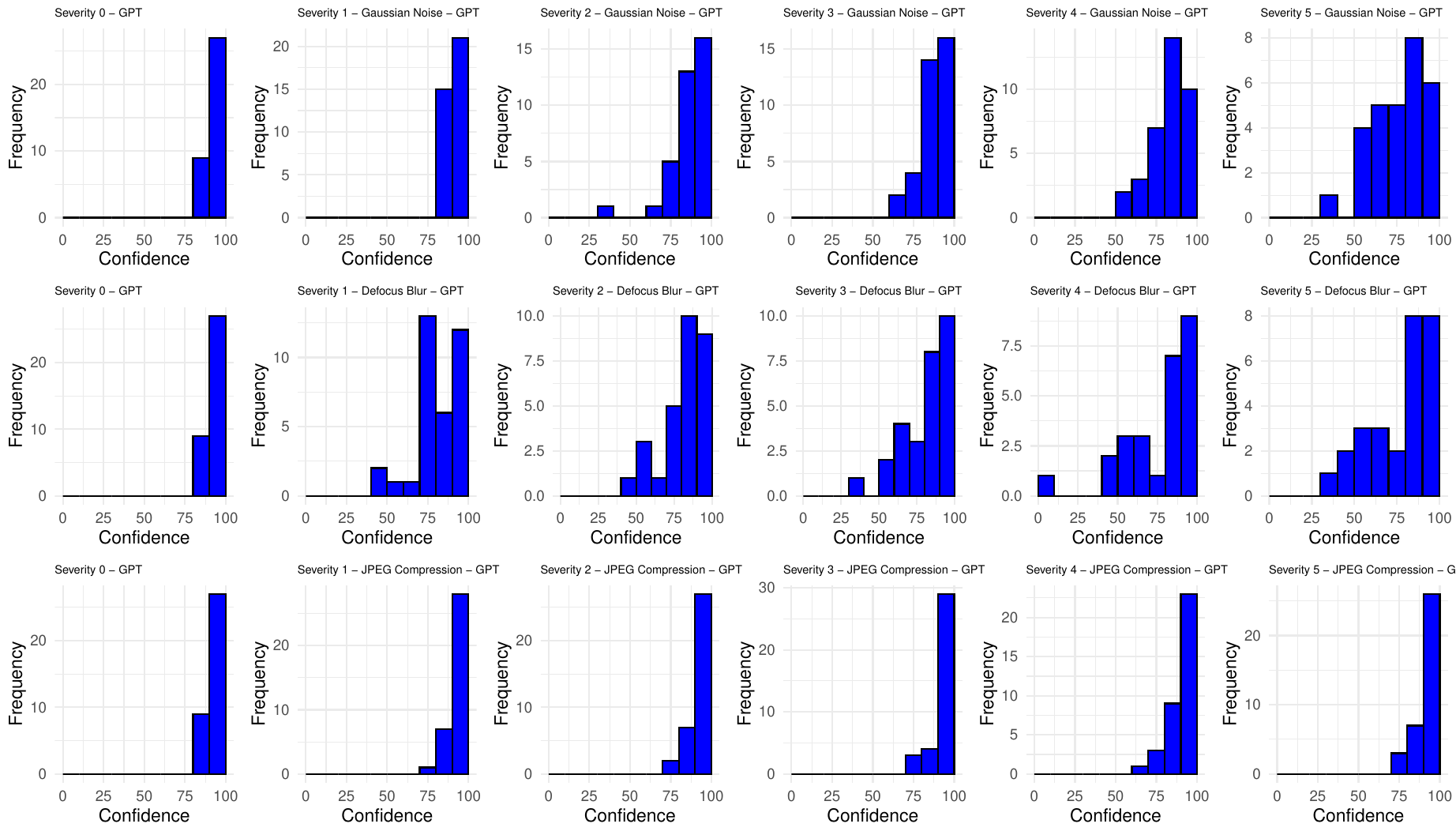}
    \caption{Confidence histograms GPT-4V easy VQA}
    \label{fig:hist_gpt}
\end{figure}

\begin{figure}[h]
    \centering
    \includegraphics[width=0.75\linewidth]{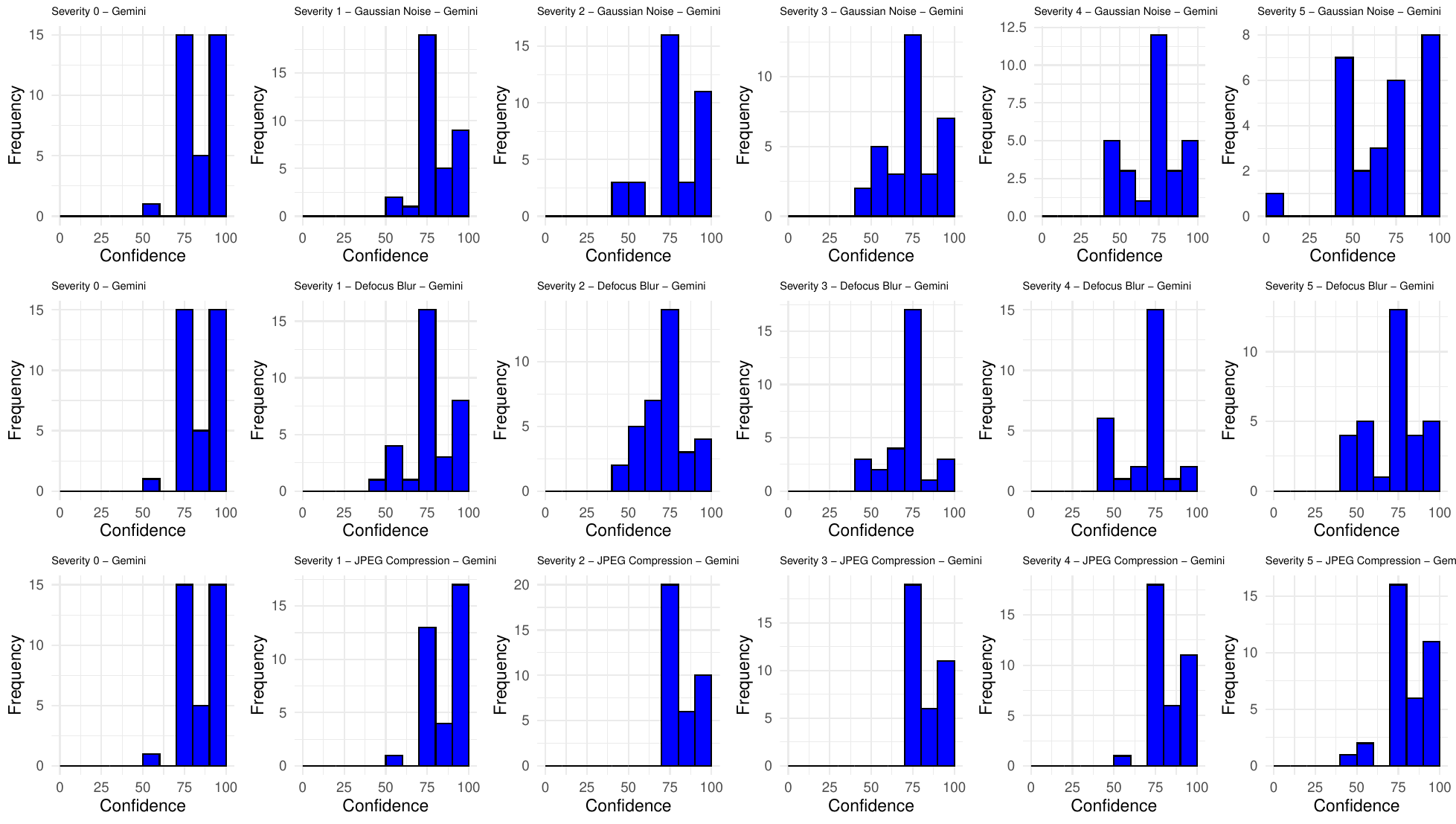}
    \caption{Confidence histograms Gemini easy VQA}
    \label{fig:hist_gemini}
\end{figure}

\begin{figure}[h]
    \centering
    \includegraphics[width=0.75\linewidth]{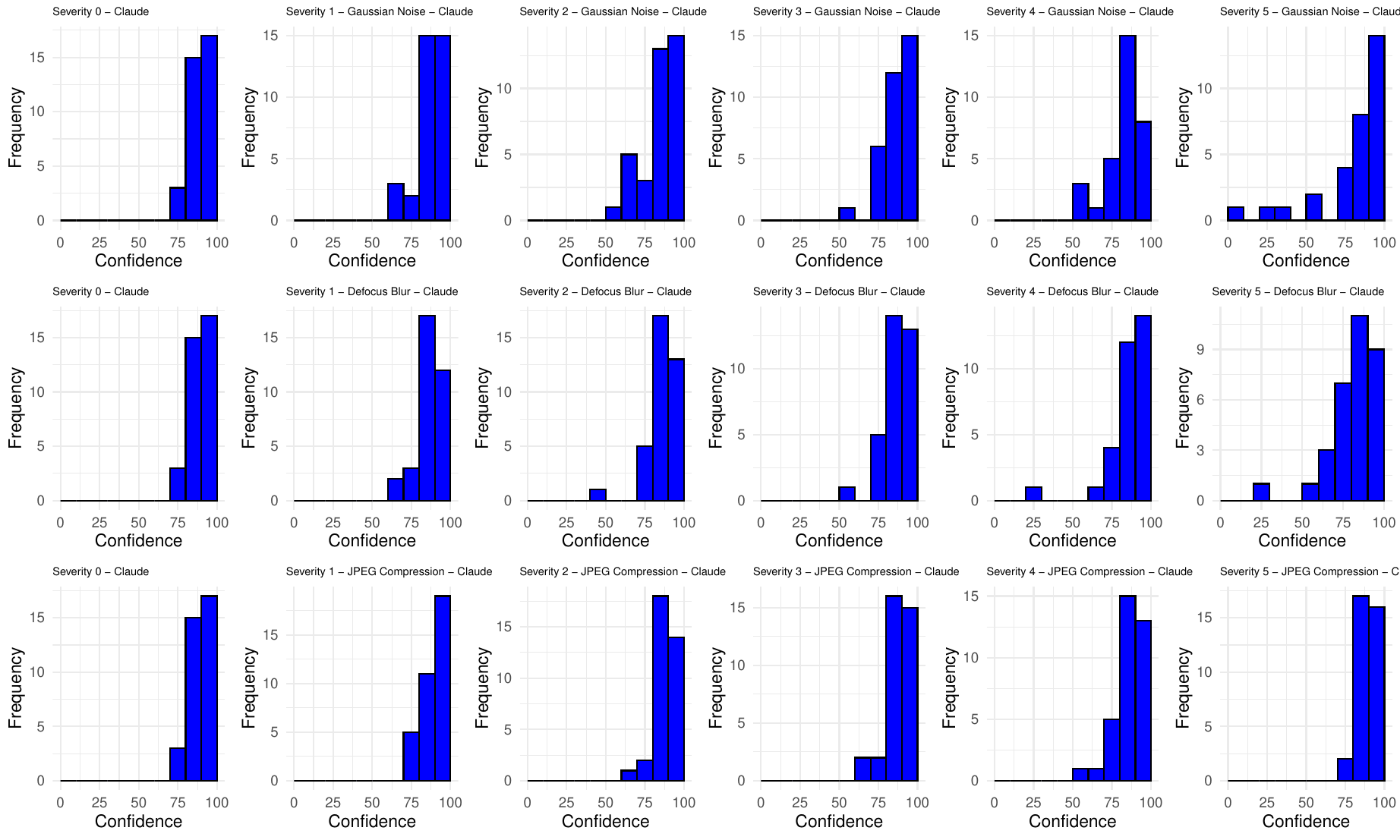}
    \caption{Confidence histograms Claude easy VQA}
    \label{fig:hist_claude}
\end{figure}

\begin{figure}[h]
    \centering
    \includegraphics[width=0.75\linewidth]{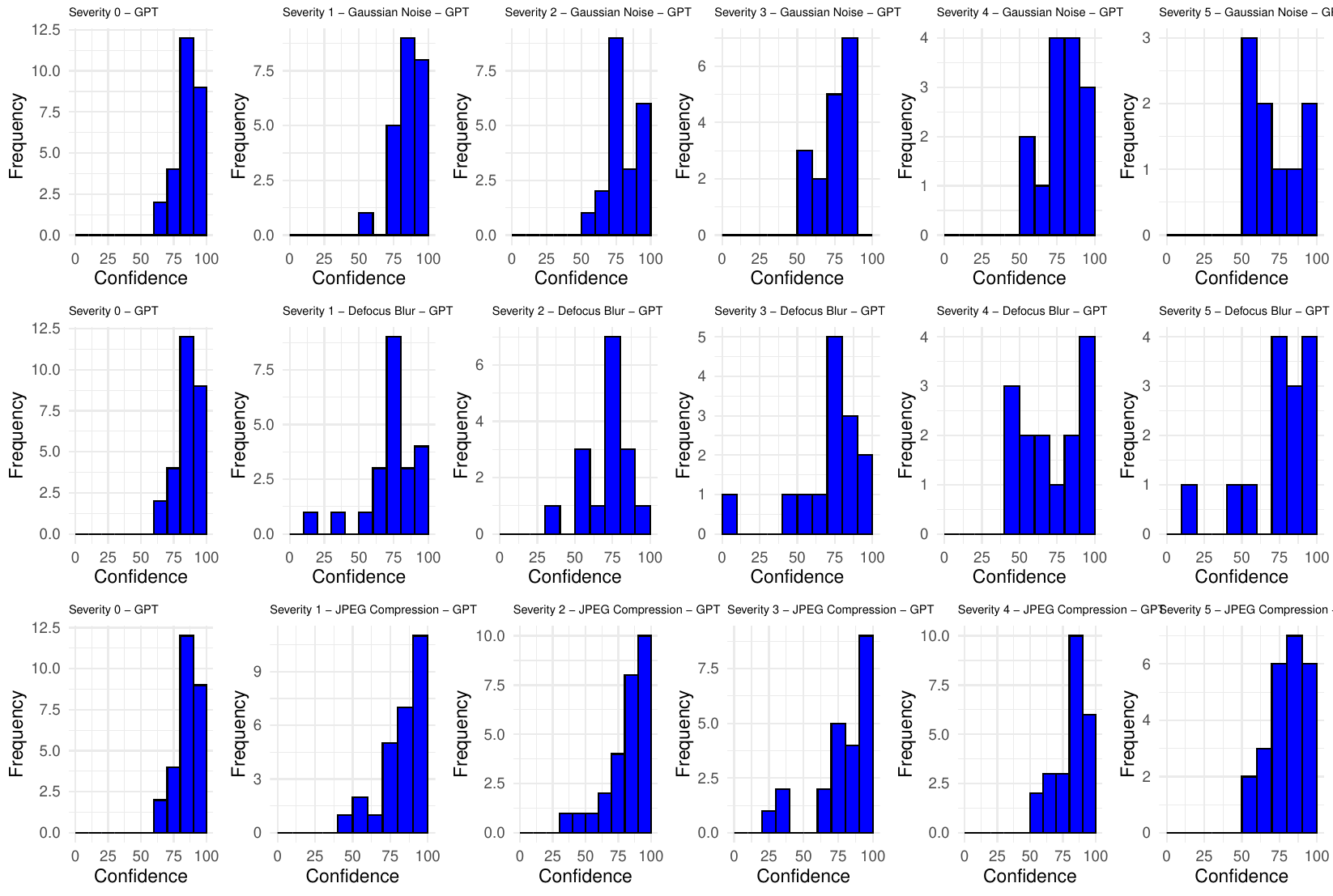}
    \caption{Confidence histograms GPT-4V hard VQA}
    \label{fig:hist_gpt_hard}
\end{figure}

\begin{figure}[h]
    \centering
    \includegraphics[width=0.75\linewidth]{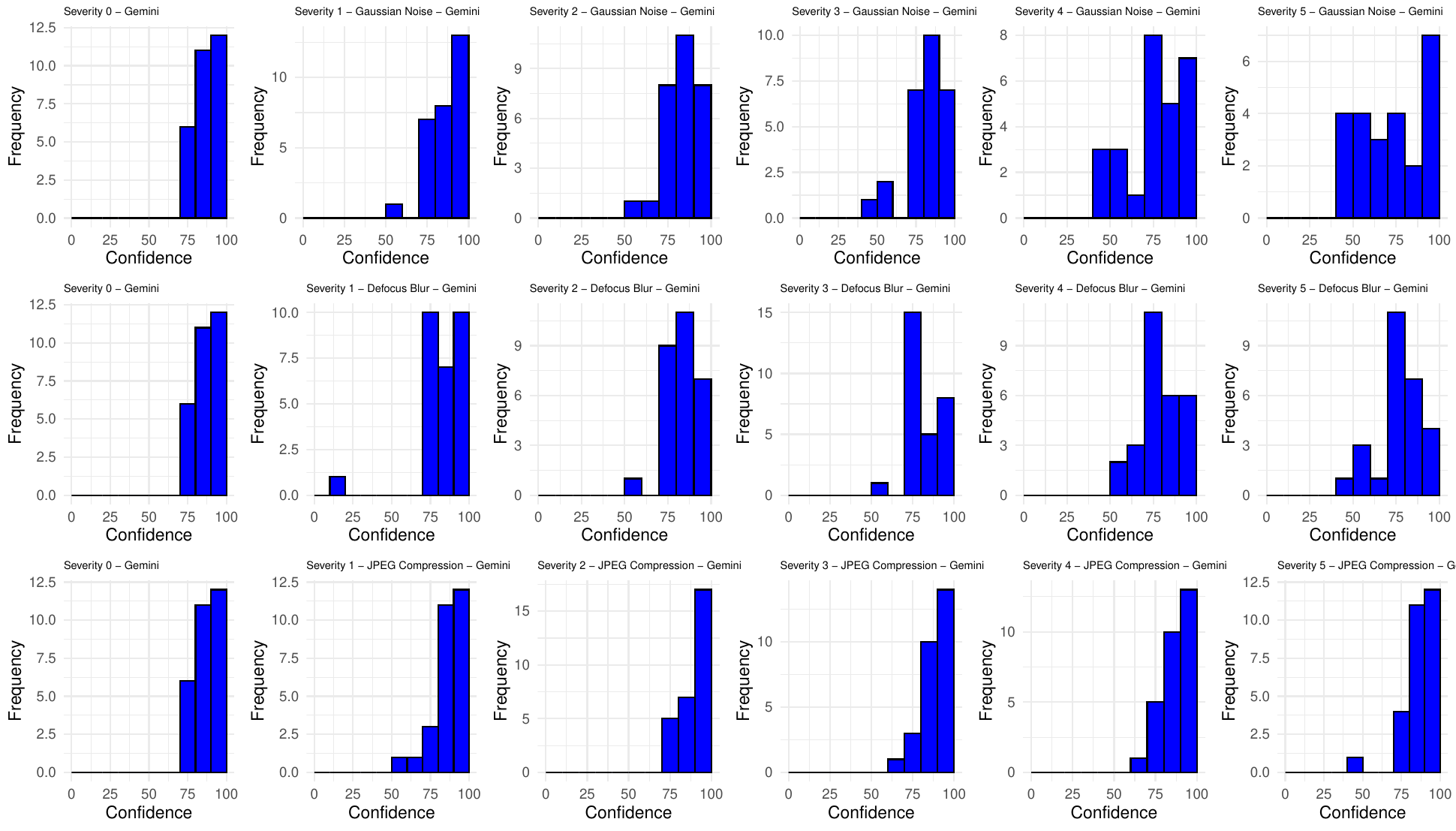}
    \caption{Confidence histograms Gemini hard VQA}
    \label{fig:hist_gemini_hard}
\end{figure}

\begin{figure}[h]
    \centering
    \includegraphics[width=0.75\linewidth]{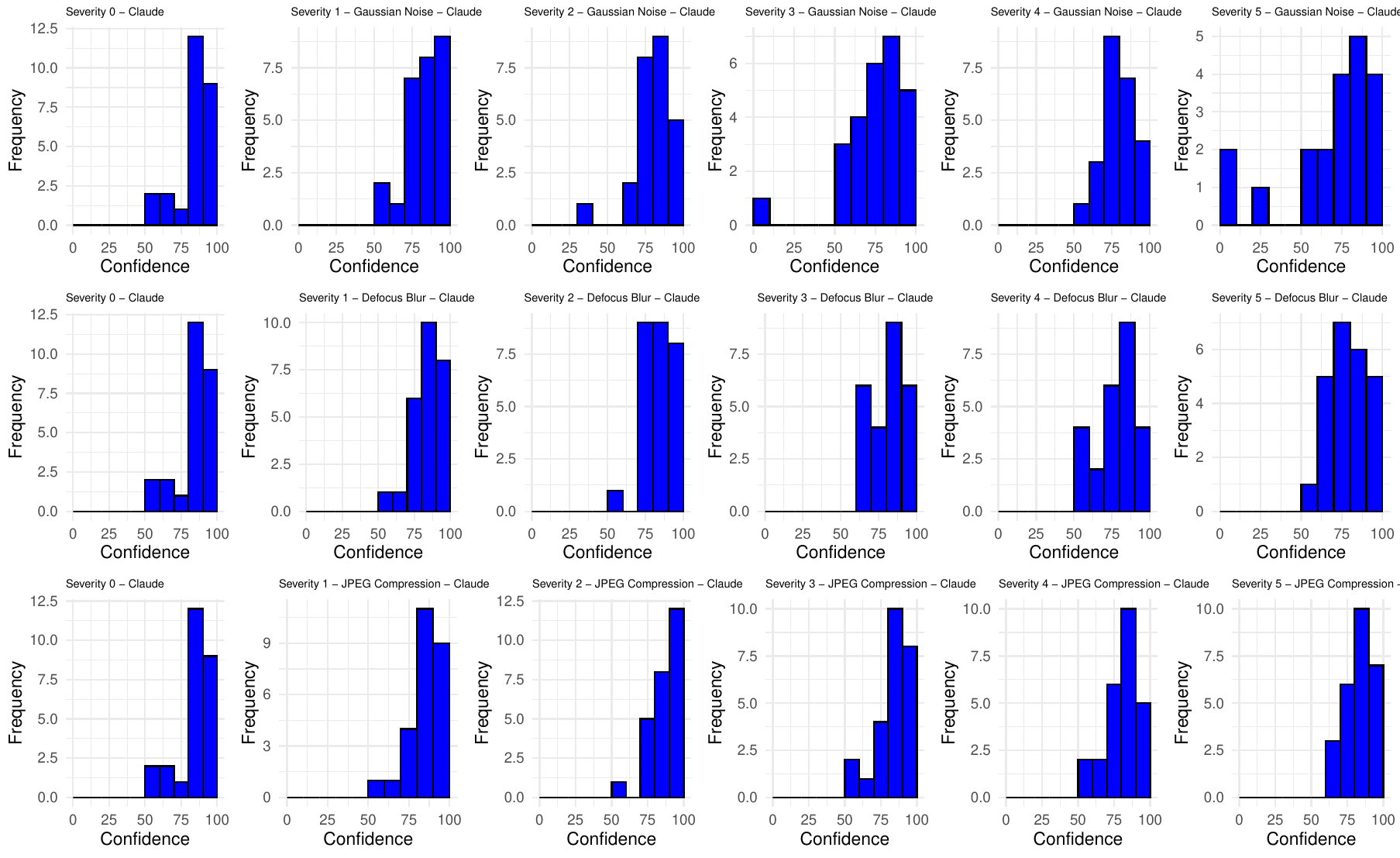}
    \caption{Confidence histograms Claude hard VQA}
    \label{fig:hist_claude_hard}
\end{figure}
\newpage
\onecolumn

\section{Responses under noise}\label{ap:appendixE}

In Table \ref{tab:claude_answers} we show how Claude even identifies the exact bridge in Kyoto but as the severity of the blur increases, it gives incorrect or more and more general answers instead of refusing to answer. 

\begin{table}[h]
    \caption{Claude answers with defocus blur\\
             \textbf{Question}: Where was this photo taken? \\
             \textbf{Correct answer}: Japan, Kyoto, Arashiyama Area, the Bridge is named Togetsu-kyo Bridge (or Toei Bridge)}
    \label{tab:claude_answers}
    \begin{tabular}{p{2.2cm}p{2.2cm}p{2.2cm}p{2.2cm}p{2.2cm}p{2.2cm}}
      \toprule
      Severity 0 & Severity 1 & Severity 2 & Severity 3 & Severity 4 & Severity 5 \\
      \midrule
      \includegraphics[width=\linewidth]{img/bridge_original.jpg} & \includegraphics[width=\linewidth]{img/bridge_defocus_blur_severity_1.jpg} & \includegraphics[width=\linewidth]{img/bridge_defocus_blur_severity_2.jpg} & \includegraphics[width=\linewidth]{img/bridge_defocus_blur_severity_3.jpg} & \includegraphics[width=\linewidth]{img/bridge_defocus_blur_severity_4.jpg} & \includegraphics[width=\linewidth]{img/bridge_defocus_blur_severity_5.jpg}\\
      \midrule
      Based on the wooden bridge structure spanning the water and the mountains in the background, this photo appears to have been taken in Kyoto, Japan at the famous Togetsukyo Bridge in Arashiyama (confidence 90\%). & Kyoto, Japan (80\%) & Kyoto, Japan (75\%) & Based on the mountainous landscape and the traditional wooden houses along the riverbank, this photo appears to have been taken in Kyoto, Japan (70\% confidence). & Based on the mountainous landscape and the body of water, which appears to be a lake, this photo was likely taken at Lake Como in Italy (75\% confidence). & Lake (60\%) \\
      \bottomrule
    \end{tabular}
\end{table}

In Table \ref{tab:gpt_answers} we show responses from GPT-4V. It first answers the question correctly and then as the severity increases, gets confused with the added noise and in the end it refuses to answer.

\begin{table}[h]
    \caption{GPT-4V answers with Gaussian noise. \textbf{Question}: What is on the sheep? \textbf{Correct answer}: An eartag}
    \label{tab:gpt_answers}
    \begin{tabular}{p{2.2cm}p{2.2cm}p{2.2cm}p{2.2cm}p{2.2cm}p{2.2cm}}
      \toprule
      Severity 0 & Severity 1 & Severity 2 & Severity 3 & Severity 4 & Severity 5 \\
      \midrule
      \includegraphics[width=\linewidth]{img/sheep_original.jpg} & \includegraphics[width=\linewidth]{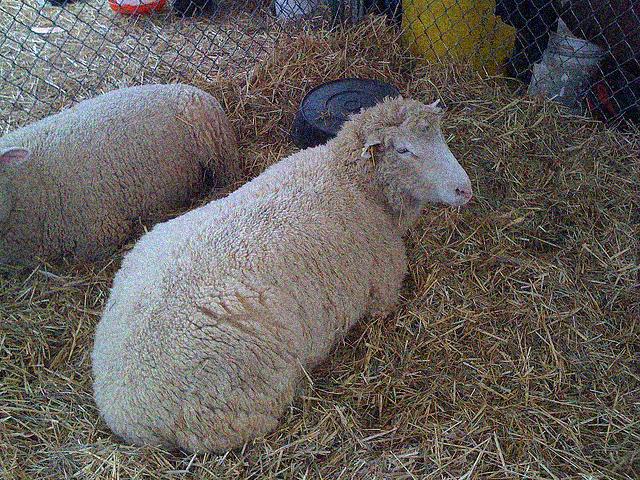} & \includegraphics[width=\linewidth]{img/sheep_gaussian_noise_severity_2.jpg} & \includegraphics[width=\linewidth]{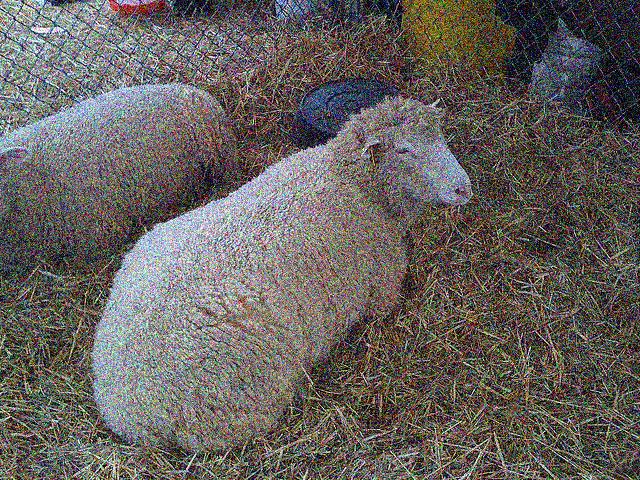} & \includegraphics[width=\linewidth]{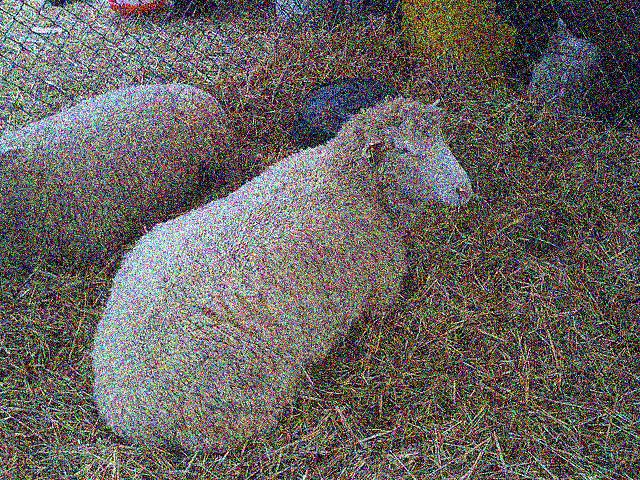} & \includegraphics[width=\linewidth]{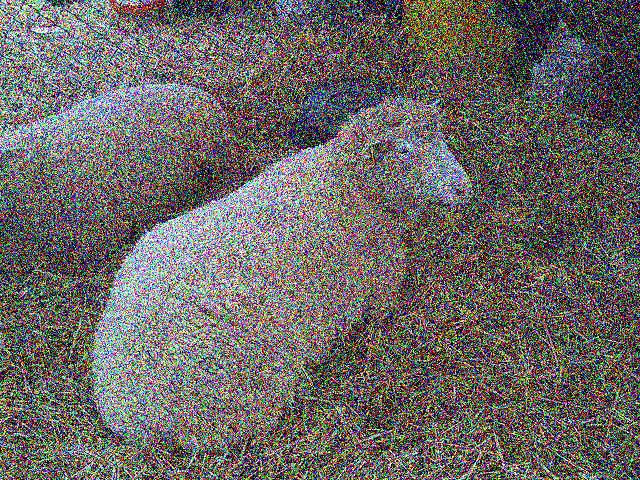}\\
      \midrule
      Tag on ear (95\%) & There is an ear tag on the sheep (confidence 100\%). & There is nothing unusual on the sheep; it appears to be just wool-covered, typical for a sheep. Answer (confidence 95\%). & There is nothing unusual on the sheep; it looks like a typical sheep with its wool coat. Answer (100\%). & The sheep is wearing a blue blanket or cover (90\% confidence). & It is difficult to accurately discern specifics about the sheep due to the distorted and noisy nature of the image. The image quality suggests it may have been altered or degraded, mostly showing pixelated and multicolored noise patterns rather than clear objects or textures. Thus, I cannot definitively make out detailed attributes such as what might be on the sheep. (Confidence: 30\%) \\
      \bottomrule
    \end{tabular}
\end{table}

\end{document}